\documentclass[letterpaper]{article} 
\usepackage[draft]{aaai2026}  
\usepackage{times}  
\usepackage{helvet}  
\usepackage{courier}  
\usepackage[hyphens]{url}  
\usepackage{graphicx} 
\usepackage{natbib}  
\usepackage{caption} 
\usepackage{algorithm}

\usepackage{newfloat}
\usepackage{listings}
\DeclareCaptionStyle{ruled}{labelfont=normalfont,labelsep=colon,strut=off} 
\floatstyle{ruled}
\newfloat{listing}{tb}{lst}{}
\floatname{listing}{Listing}
\usepackage{algpseudocode}
\algrenewcommand\algorithmicindent{1em}
\usepackage{amsmath}
\usepackage{amssymb}
\usepackage{graphicx}
\usepackage{xcolor}
\usepackage{bm}
\usepackage{subfigure}

\newcommand{\model}{ProToM}

\title{\model{}: Promoting Prosocial Behaviour via Theory of Mind-Informed Feedback}

\author {
    Matteo Bortoletto\textsuperscript{\rm 1,\S},
    Yichao Zhou\textsuperscript{\rm 2,\S}, 
    Lance Ying\textsuperscript{\rm 3,4},
    Tianmin Shu\textsuperscript{\rm 5},
    Andreas Bulling\textsuperscript{\rm 1}
}
\affiliations {
    \textsuperscript{\rm 1}
    University of Stuttgart \quad
    \textsuperscript{\rm 2}
    Peking University \quad
    \textsuperscript{\rm 3}
    Harvard University \\
    \textsuperscript{\rm 4}
    Massachusetts Institute of Technology \quad
    \textsuperscript{\rm 5}
    Johns Hopkins University \\ 
    \textsuperscript{\rm \S}
    Work done while at Johns Hopkins University \\ 
    \hspace{1em} \\
    Correspondence to: matteo.bortoletto@vis.uni-stuttgart.de
}

\begin{document}

\maketitle

\begin{abstract}
    While humans are inherently social creatures, the challenge of identifying when and how to assist and collaborate with others -- particularly when pursuing independent goals -- can hinder cooperation.
To address this challenge, we aim to develop an AI system that provides useful feedback to promote prosocial behaviour -- actions that benefit others, even when not directly aligned with one's own goals.
We introduce \textit{\model{}}, a Theory of Mind-informed facilitator that promotes prosocial actions in multi-agent systems by providing targeted, context-sensitive feedback to individual agents. 
\model{} first infers agents' goals using Bayesian inverse planning, then selects feedback to communicate by maximising expected utility, conditioned on the inferred goal distribution.
We evaluate our approach against baselines in two multi-agent environments: Doors, Keys, and Gems, as well as Overcooked. 
Our results suggest that state-of-the-art large language and reasoning models fall short of communicating feedback that is both contextually grounded and well-timed -- leading to higher communication overhead and task speedup.
In contrast, \model{} provides targeted and helpful feedback, achieving a higher success rate, shorter task completion times, and is consistently preferred by human users.

\end{abstract}

\section{Introduction}

Prosocial behaviour is a broad class of actions that benefit other individuals \cite{kakulte2023prosocial} or society as a whole \cite{carattini2023trust}.
However, humans do not always engage in prosocial behaviour because they may not know if others need help, or how they can help.
As AI systems increasingly mediate human behaviour in everyday life, they inevitably shape not only individual outcomes but the broader social fabric.  
Therefore, while much of the discussion around responsible AI focuses on preventing harm, an equally important goal is to actively promote social good, e.g., encouraging courtesy, reciprocity, or equitable resource sharing  \cite{binns2018fairness}.    

\begin{figure}[t]
    \centering
    \includegraphics[width=\linewidth]{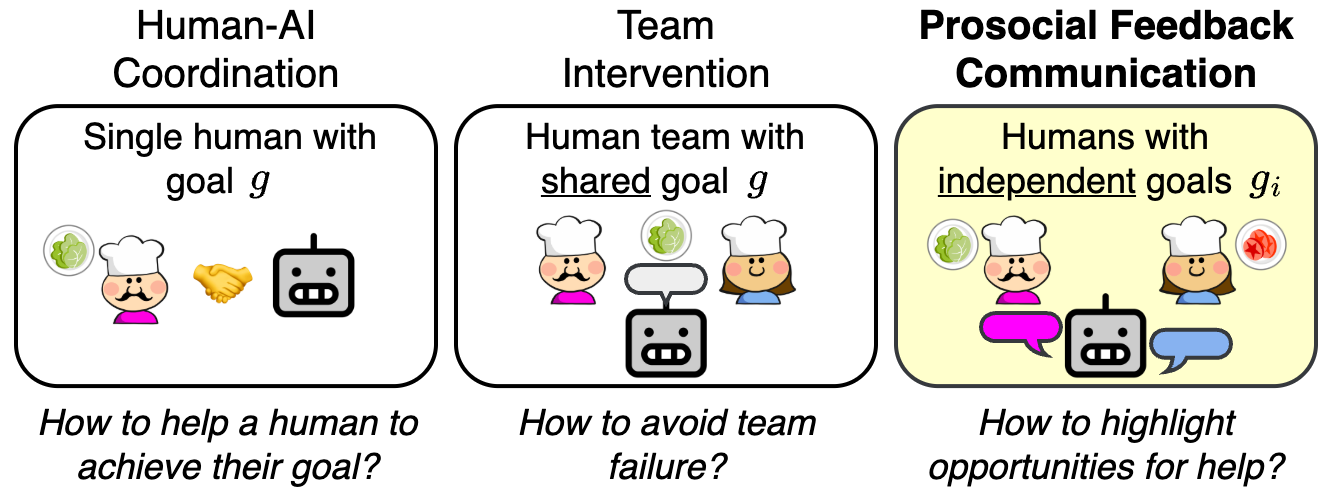}
    \caption{In contrast to the dominant paradigms of human-AI coordination and team intervention, we aim to encourage prosocial interactions among human agents pursuing independent goals. We introduce \model{}, a \textit{facilitator} that promotes prosocial actions by providing targeted, context-sensitive feedback to individual agents.}
    \label{fig:teaser}
\end{figure}

In this work, we argue that a promising yet under-explored direction is to leverage AI systems to encourage prosocial interactions among humans. 
This represents a shift from the dominant paradigm of human-AI coordination (Figure~\ref{fig:teaser}, left), where an AI assistant and a human operate as a joint unit with a shared goal \cite{carroll2019utility, puig2023nopa, ying2024goma, zhi2024pragmatic, chang2024partnr} or to team intervention (Figure~\ref{fig:teaser}, middle) where the AI assists a team with a shared goal \cite{seo2023automated, zhang2024risk}. 
By contrast, our proposed \textbf{prosocial facilitator paradigm} introduces an AI that operates as an observer and \textit{facilitator} among multiple humans that pursue different goals (Figure~\ref{fig:teaser}, right). 
Our key research question is:
\begin{quote}
    \emph{Can we design an AI facilitator that identifies opportunities for prosocial behaviour among humans and provides timely, helpful feedback to encourage it?}
\end{quote}
Designing such a facilitator presents substantial challenges, as choosing when and what feedback to communicate highly depends both on the state of the environment and the agents' internal states, such as beliefs and goals, which must be inferred only from available observations. 
Compared to human-AI coordination, an additional challenge arises from the need to model multiple agents, each with their own beliefs and goals, as well as how they interact with the environment and with one another based on those internal states.

We introduce \model{}\footnote{Project page: \url{https://git.hcics.simtech.uni-stuttgart.de/public-projects/ProToM}} -- a Theory of Mind-informed facilitator that observes agents in an environment and provides feedback in real-time to encourage prosocial behaviour. 
\model{} operates by first inferring the distribution over possible goals of each agent via Bayesian inverse planning. 
Conditioned on the inferred goal distributions and the current state of the environment, it then evaluates a set of possible feedback messages by computing their expected utility -- measuring how helpful each feedback would be in guiding the agent toward prosocial actions that benefit another agent. 
It then selects the feedback with the highest expected utility and communicates it to the target agent only if the divergence between the agent's simulated plan with or without communicating the feedback is large enough, i.e., if the agent is expected to behave differently in the absence of feedback.
This ensures that \model{} delivers feedback only when it is both relevant and likely to be effective, resulting in timely and useful communication.
Together with the feedback message, \model{} provides an explanation based on the inferred agents' goals, helping the recipient understand not just what to do, but why the feedback is relevant and why it should be executed in a particular context.

We first conduct experiments with simulated human agents on two common multi-agent environments, Multi-Agent Doors, Keys, and Gems \cite{zhi2024pragmatic} and Overcooked \cite{carroll2019utility}. 
Our results show that state-of-the-art large vision-language and reasoning models struggle to deliver meaningful feedback and tend to increase communication overhead.
In contrast, \model{} consistently provides useful and well-timed feedback, resulting in a perfect task success rate, faster task completion, and reduced communication overhead. 
We further conducted a human study with real human participants. 
The results show that participants perceived \model{} to have a stronger understanding of their own goals and to provide feedback that is more helpful, appropriate in both content and amount, and better explained. 

In summary, our main contribution includes: (1) a new human-AI  paradigm for prosocial feedback communication; (2) a method, \model{}, that enables an AI facilitator to jointly infer agents' mental states and generate ToM-informed feedback to promote prosocial behaviour among them; (3) a human study that evaluates the performance of models in real-time assistance with human participants, and humans' perception of them. 

\section{Problem Formulation}

We consider an AI facilitator as an omniscient observer that provides feedback to $n$ interacting agents with the goal of promoting prosocial behaviour. 
In the most general case, this setting can be formalised as a two-level POMDP: inner agents play a partially‑observable game with their own goals, while the outer assistant plays a POMDP whose hidden component is the observed agents' internal state and whose only available action is the feedback communication.
Formally, we can formulate the problem as a tuple
\begin{equation}
    \langle
    S,
    \{A^i\}_{i=1}^{n},
    \mathcal{F},
    T,
    \{\Omega^i\}_{i=1}^{n},
    O,
    \{G^i\}_{i=1}^{n}
  \rangle
  \label{eq:afg_tuple}
\end{equation}
where: 
$S$ is the state space, 
$A^i$ is the action set of agent $i\in\{1,\dots,n\}$, 
$\mathcal{F}$ is the finite set of \emph{feedback messages} the assistant can communicate, 
$T(s'\mid s,\bm{a})$ is the state‑transition kernel given joint action $\bm{a}=(a^1,\dots,a^n)\in A^1\times\dots\times A^n$,
$\Omega_i$ is the observation space of agent $i$, 
$O(o^i\mid s,f)$ is the observation kernel, 
and $G^i$ is the set of possible goals of agent $i$. 
The feedback $f\in \mathcal{F}$ is included in the observation kernel $O$, since feedback may convey extra information about the state that agents cannot directly observe.

\paragraph{Agent Model} 
In partially observable environments, since agents do not know the true state $s_t$, they maintain a probability distribution over the possible states, called the \textit{belief state} $b_t$. 
The probabilistic generative model for agents' behaviour is defined as follows:
\begin{alignat}{3}
&\textit{Goal Prior:} \quad &g^i &\sim P(g^i) \label{eq:goal-prior} \\
&\textit{Belief Update:} \quad &b^i_t &\sim P(b^i_t \mid o^i_t) \label{eq:belief-update} \\
&\textit{Action Selection:} \quad &a^i_t &\sim P(a^i_t \mid b^i_t, g^i) \label{eq:action-selection} \\
&\textit{State Transition:} \quad &s_{t+1} &\sim T(s_{t+1} \mid s_t, a^i_t, \bm{a}^{-i}_t) \label{eq:state-transition}
\end{alignat}
where $\bm{a}^{-i}_t$ indicates other agent's actions.
To model planning, we assume that each agent selects actions $a^i_t$ according to a policy $\pi^i(a^i_t \mid b^i_t, g^i)$, which may reflect approximate rationality (e.g., Boltzmann-rational planning) or be derived from heuristic or rule-based strategies. 
In fully observable settings, $b_t^i=s_t$. 

\paragraph{Facilitator Model} 
Given an observed trajectory $\tau = \{(s_t, \bm{a}_t)\}_{t=1}^T$, the observer AI facilitator must decide whether to provide feedback, and if so, which $f \in \mathcal{F}$ to communicate: $f \sim P(f \mid \tau)$.
Since the agents' internal state is not visible to the AI facilitator, the environment also is a POMDP from the facilitator's perspective.
\section{Method}

\begin{figure*}[t]
    \centering
    \includegraphics[width=0.9\linewidth]{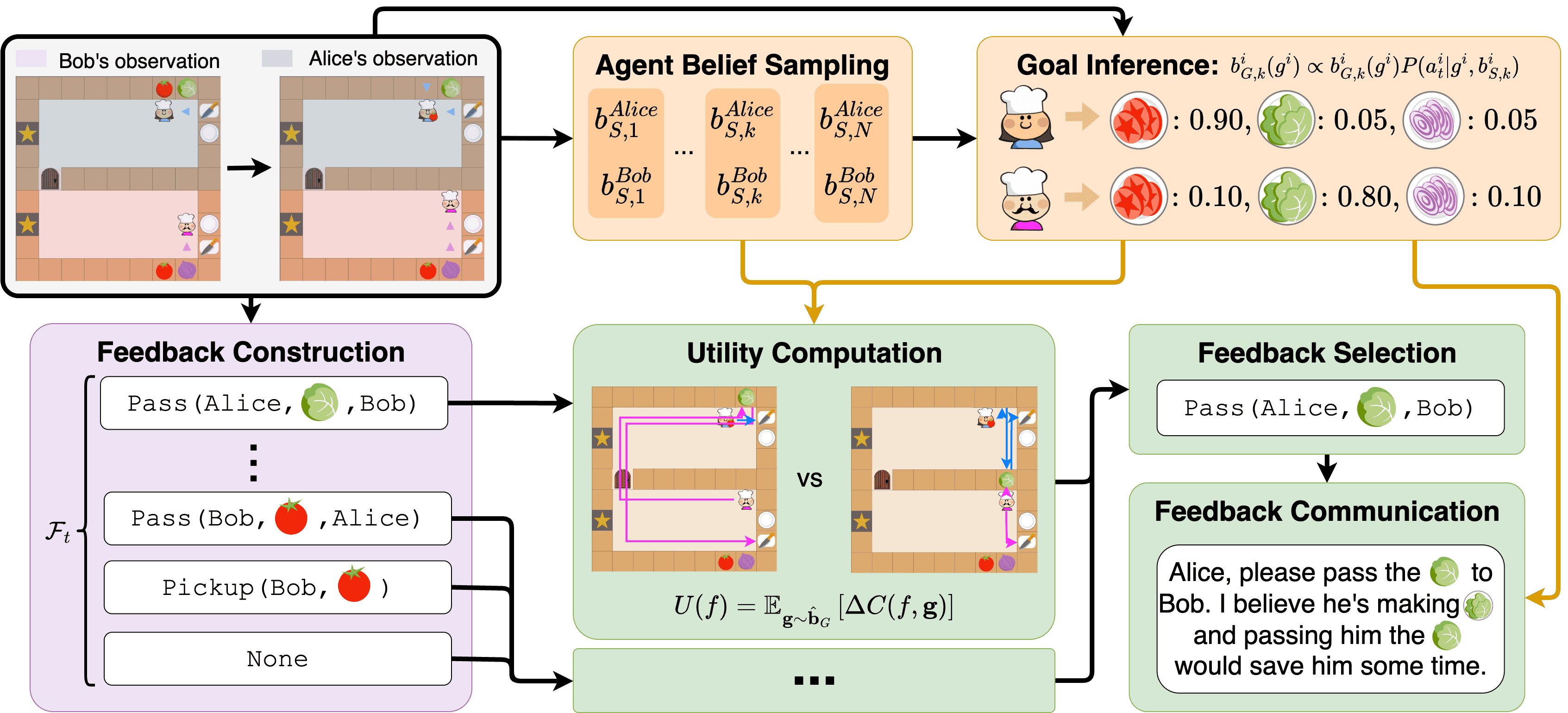}
    \caption{Overview of \model{}, a Theory of Mind-informed facilitator that promotes prosocial behaviour by communicating real-time feedback to agents pursuing independent goals. \model{} observes agents acting in a shared environment, infers their goal distributions via Bayesian inverse planning, and evaluates possible feedback messages based on expected utility. Feedback is paired with an explanation grounded in the inferred goals to clarify why the suggestion is relevant and helpful.}
    \label{fig:model}
\end{figure*}

\model{} actively assists both agents in an environment by observing their actions, inferring their goals, and providing feedback when it is expected to improve prosocial behaviour and thus overall performance. 
We present an overview of the method in Figure~\ref{fig:model} and Algorithm~\ref{alg:model}.

\begin{algorithm}[t]
\caption{\model{}}
\begin{algorithmic}[1]
\Require Goal set $\mathcal{G}=G^1\times\cdots\times G^n$, planners $\boldsymbol{\pi} = (\pi^1, \cdots, \pi^n)$, performance metric $\ell(\cdot)$, thresholds $\phi$, $\epsilon$
\State \textbf{Initialise}: belief particles $\{\{(b_{S, k}^i, b_{G, k}^i)\}_{k=1}^N\}_{i=1}^n$ based on initial state $s_0$, $t \gets 1$
\While {$t = T_{max}$ or Done}
    \State Observe state $s_t$ and agent action $\bm{a}_t=(a^1_t,\cdots, a^n_t)$
    \For{agent $i \in \{1, \cdots, n\}$}
        \State Get agent observation $o_t^i$ from $s_t$
        \For{belief particle $k \in \{1, \cdots, N\}$}
            \State $b_{S, k}^i \gets \textsc{BeliefUpdate}(b_{S, k}^i, o_t^i,\bm{a}_t)$ 
            \State $b_{G, k}^i \gets \textsc{GoalUpdate}(b_{G, k}^i, b_{S, k}^i, G^i, a_t^i)$
        \EndFor
        \State $\hat{b}_G^i \gets \frac{1}{N}\sum_{k=1}^N b_{G, k}^i$
    \EndFor
        \State $\mathcal{F}_t \gets \textsc{ConstructFeedback}(s_t)$
    \State $U(f) \gets \textsc{ComputeUtility}(s_t, \mathcal{G}, \hat{\mathbf{b}}_G, \mathcal{F}_t, \boldsymbol{\pi}, \ell)$
    \State $\hat{\mathcal{F}_t}\gets\{f\in\mathcal{F}_t\mid U(f) = \max_{f'\in\mathcal{F}_t}U(f')>\phi, \mathrm{div}(f)>\epsilon\}$ 
    \If {$\hat{\mathcal{F}_t} \ne \emptyset$}
        \State $\hat{f}\sim \mathrm{Uniform}(\hat{\mathcal{F}_t})$
        \State $e \gets \textsc{Explanation}(\mathcal{G}, \hat{\mathbf{b}}_G, s_t, \hat{f})$
        \State Communicate feedback $\hat{f}$ with explanation $e$
    \EndIf
    \State $t\gets t+1$
\EndWhile
\end{algorithmic}
\label{alg:model}
\end{algorithm}
\begin{algorithm}[t]
\caption{\textsc{ComputeUtility}}
\begin{algorithmic}[1]
\Require Current state $s_t$, goal set $\mathcal{G}=G^1\times\cdots\times G^n$, inferred goal probability distributions $\hat{\mathbf{b}}_G = (\hat{b}^1_G, \cdots, \hat{b}^n_G)$, candidate feedback set $\mathcal{F}_t$, planners $\boldsymbol{\pi} = (\pi^1, \cdots, \pi^n)$, performance metric $\ell(\cdot)$
\State Initialise $U(f) \gets 0$ for all $f \in \mathcal{F}_t$
\For{each $\mathbf{g} = (g^1, \cdots, g^n)\in \mathcal{G}$} 
    \State $C(\varnothing, \mathbf{g}) \gets \ell(\boldsymbol{\pi} \mid \mathbf{g})$
    \For{each feedback $f \in \mathcal{F}_t$}
        \State $C(f, \mathbf{g}) \gets \ell(\boldsymbol{\pi} \mid f, \mathbf{g})$
        \State $\Delta C \gets C(\varnothing, \mathbf{g}) - C(f, \mathbf{g})$
        \State $U(f) \gets U(f) + \Delta C \cdot \prod_{i=1}^n \hat{b}_G^i(g^i)$
    \EndFor
\EndFor
\end{algorithmic}
\label{alg:utility}
\end{algorithm}

\subsection{Maintaining Beliefs About Agents}
To simulate and interpret the behaviour of agents, \model{} maintains a dynamic belief distribution $b_F^i$ for each agent $i$, capturing both their internal belief about the environment and their underlying goal. This belief distribution is approximated using a particle filter with $N$ particles per agent:
\begin{align}
    b_F^i = \{(b_{S, k}^i, b_{G, k}^i)\}_{k=1}^N,
\end{align}
where each particle $k$ contains a sampled belief state $b_{S, k}^i$ representing agent $i$'s internal model of the environment, and a probability distribution $b_{G, k}^i$ over possible goals for agent $i$, computed via Bayesian inverse planning.

\paragraph{Agent Belief Sampling} 
While \model{} has access to the full environment state, it must infer agents' goals from their actions under partial observability. 
To do this, for each particle $k \in \{1, \dots, N\}$, a sample of the agent $i$'s belief state is drawn from a distribution conditioned on the agent's current observation $o^i_t$:
\begin{align}
    b_{S, k}^i \sim P(b^i_t \mid o^i_t).
\end{align}
At each timestep $t$, these sampled beliefs are updated using the environment’s state transition dynamics. 
To ensure that particles remain consistent with the agent's actual observations, \model{} resamples any particle $b_{S, k}^i$ that becomes incompatible with the latest observation $o_t^i$. 
This ensures that the facilitator's estimate of the agents' beliefs are consistent with what the agent could reasonably believe based on their observations.

\paragraph{Goal Inference} 
Given a sampled belief state $b_{S,k}^i$, \model{} infers the agent's goal using Bayesian inverse planning. 
Each particle maintains a goal distribution $b_{G,k}^i(g^i)$, which is updated by conditioning on the observed action $a_t^i$ and the particle's belief: 
\begin{align}
    b_{G,k}^i(g^i) \propto b_{G,k}^i(g^i) \cdot P(a_t^i \mid g^i, b_{S, k}^i).
\end{align}
This update reflects the likelihood of the agent taking action $a_t^i$ under goal $g^i$, assuming it holds belief $b_{S,k}^i$.
\model{}'s overall belief about agent $i$'s goal is then computed by averaging across all particles:
\begin{align}
    \hat{b}_G^i(g^i) = \frac{1}{N} \sum_{k=1}^N b_{G,k}^i(g^i).
\end{align}

\subsection{Feedback Selection}
Given a finite set of candidate feedback messages, \model{} performs feedback selection in three steps. 
First, it evaluates each candidate feedback message by computing its expected utility, conditioned on the inferred goals and belief states of the agents. 
Second, \model{} decides whether to issue new feedback by comparing the expected utility of the best candidate against a threshold and estimating the divergence between the agent’s predicted behaviour with and without the feedback. 
This process ensures that feedback is only given when it is likely to meaningfully improve the agent's performance. 
Finally, if a feedback message is selected, \model{} generates a corresponding explanation to help the agent understand the reasoning behind the suggestion. 
This process is repeated at each timestep $t$, unless the agents are already executing feedback. 

\paragraph{Feedback Construction}
The set of candidate feedback messages $\mathcal{F}_t$ is constructed based on the current full environment state $s_t$, including object locations, agent positions, and affordances. 
This ensures that all feedback options are feasible. 
For example, if an agent cannot physically reach a particular item (e.g., due to immovable obstacles), we exclude feedback that would suggest picking up or passing that item. 

\paragraph{Utility Computation}
Our utility function evaluates whether the feedback promotes prosocial actions that improve the overall task efficiency under sampled goal hypotheses from \model{}'s belief particles for all $n$ agents (Algorithm~\ref{alg:utility}). 
We denote these samples as:
\begin{align}
    \mathbf{g} \sim \hat{\mathbf{b}}_G
\end{align}
where
\begin{align}
    \mathbf{g} = (g^1, g^2, \cdots, g^n), \quad
    \hat{\mathbf{b}}_G = (\hat{b}_G^1, \hat{b}_G^2, \cdots, \hat{b}_G^n).
\end{align}
The utility of a feedback message $f \in \mathcal{F}_t$ under $\mathbf{g}$ is defined as:
\begin{align}
    U(f)=\mathbb{E}_{\mathbf{g} \sim \hat{\mathbf{b}}_G} \left[\Delta C(f,\mathbf{g}) \right] 
    \label{eq:utility}
\end{align}
where we define the marginal improvement over not providing any feedback as:
\begin{align}
    \Delta C(f,\mathbf{g}) &= C(\varnothing,\mathbf{g}) - C(f,\mathbf{g}).
\end{align}
Theoretically, $C(f,\mathbf{g})$ can be any performance metric $\ell(\cdot)$ that evaluates the agents' plans $\boldsymbol{\pi}=(\pi^1, \pi^2, \cdots, \pi^n)$ generated under $f$ and $\mathbf{g}$:
\begin{align}
    C(f,\mathbf{g}) = \ell(\boldsymbol{\pi} \mid f, \mathbf{g}). 
\end{align}
In our case, $\ell(\cdot)$ corresponds to the number of steps to achieve $\mathbf{g}$, either with or without executing $f$ beforehand.
A positive utility $U(f)$ indicates that providing feedback $f$ is expected to help agents achieve their individual goals more efficiently -- by decreasing the global number of steps required to reach task completion. 

When a feedback message is targeted to agent $i$, it directly influences the policy $\pi^i$ of that agent. 
We assume that agent $i$ will follow the feedback upon receiving it, and continue to plan accordingly until either the feedback directive is completed, or it becomes infeasible. 
After that, the agent resumes planning toward its goal $g^i$. 

\paragraph{Feedback Selection}
Among all candidates, we consider only feedback messages whose utility exceeds a threshold $\phi$ and select those with the highest utility. 
The threshold $\phi$ can be considered as the minimum expected benefit a feedback message must provide to even be considered.
To avoid over-communicating and ensure that feedback is communicated at an appropriate time without interrupting the agent for negligible discrepancies, \model{} communicates the feedback only when the divergence from the agent's expected behaviour, $\mathrm{div}(f)$, is greater than a threshold $\epsilon$:
\begin{align}
    \hat{\mathcal{F}_t}=\{f\in\mathcal{F}_t\mid U(f) = \max_{f'\in\mathcal{F}_t}U(f')>\phi, \mathrm{div}(f)>\epsilon\}.
\end{align}
If $\hat{\mathcal{F}_t}= \emptyset$, \model{} skips communication in this timestep.
If there is exactly one candidate in $\hat{\mathcal{F}_t}$, it is selected directly. 
If multiple feedback candidates satisfy these criteria, one is chosen uniformly at random: 
\begin{align}
    \hat{f} \sim \mathrm{Uniform}(\hat{\mathcal{F}_t}),\,\text{if} \,\hat{\mathcal{F}_t}\ne \emptyset.
\end{align}

\paragraph{Feedback Explanation}
\model{} provides a natural-language explanation for each selected feedback by populating predefined templates $\mathcal{T}$. 
Specifically, each explanation $e$ is determined by the agents' most probable goals in \model{}'s belief particles $\hat{\mathbf{g}}^{-i}$, the selected feedback $\hat{f}$, and the current state of the environment $s_t$:
\begin{align}
    e = \mathcal{T}(\hat{\mathbf{g}}^{-i}, \hat{f}, s_t), \,\,
    \,\, \hat{\mathbf{g}}^{-i} = \arg\max_{\mathbf{g}^{-i}} \hat{\mathbf{b}}_{G}^{-i}(\mathbf{g}^{-i}) .
\end{align}
The template highlights the inferred goals of other agents, and provides a concise reason for the feedback based on the current environment state (e.g. another agent may lack access to, or take much longer to reach a needed object), helping the recipient better understand the context and intent of the feedback.

\section{Experiments}

\subsection{Environments}
We evaluate our model against baselines in two popular multi-agent environments: Multi-Agent Doors, Keys, and Gems \cite[mDKG]{zhi2024pragmatic} and Overcooked \cite{carroll2019utility}. 
In both environments, $n=2$ agents are assigned individual goals, and a facilitator observes their actions and provides feedback to encourage prosocial behaviour. 
Importantly, we adapt these environments from their standard cooperative settings -- where agents typically work together toward a shared goal -- to a novel configuration in which each human pursues a separate objective. 
This distinction is crucial, as it allows us to investigate how the AI facilitator can promote prosocial behaviour in scenarios where agents are not explicitly required to collaborate.
For each environment, we consider scenarios where (1) feedback is not needed, as agents can optimally complete their task by themselves; (2) feedback is useful to improve task completion efficiency; (3) feedback is essential to allow one of the agents to complete their task. 

\textbf{mDKG} \cite{zhi2024pragmatic} is a fully observable, 2D grid-world where the objective is to collect coloured gems, which are often located behind locked doors. 
Unlocking a door requires an agent to possess a key that matches the door's colour. 
In addition to navigation, agents can pick up items, hand them over to other agents, or unlock doors when they possess a matching key. 
The possible feedback messages involve either unlocking a door or handing over a key. 

\begin{figure*}[t]
    \centering

    \begin{minipage}{\textwidth}
        \centering
        \subfigure{%
            \includegraphics[width=\linewidth]{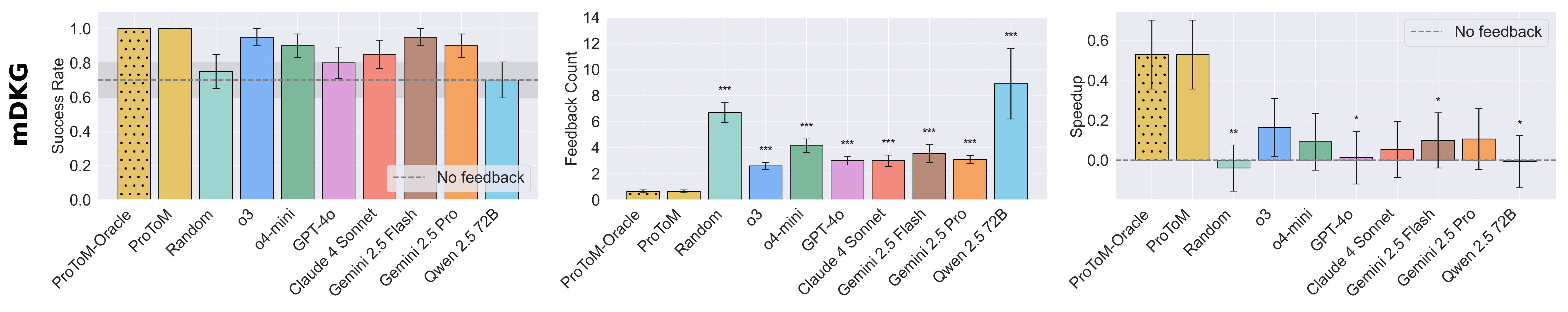}
            \label{fig:sim-mdkg}
        }
    \end{minipage}

    \begin{minipage}{\textwidth}
        \centering
        \subfigure{%
            \includegraphics[width=\linewidth]{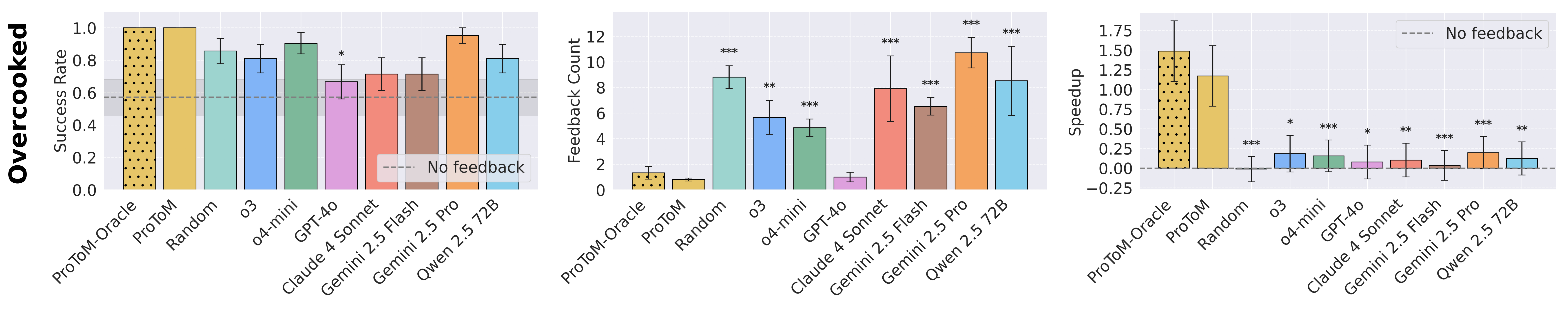}
            \label{fig:sim-oc}
        }
    \end{minipage}
    \caption{Simulation results comparing \model{} to baseline models on mDKG and Overcooked. \model{} achieves perfect success rates and significantly higher task speedup with minimal communication overhead. In contrast, baselines struggle to provide helpful, context-aware feedback and tend to over-communicate (***: $p< 0.001$, **: $p<0.01$, *: $p<0.05$). 
    }
    \label{fig:sim-results}
\end{figure*}

\textbf{Overcooked} \cite{carroll2019utility} is a 2D grid-based kitchen in which agents prepare and deliver meals. 
Agents can pick up and put down items, carry food to knife stations for chopping, and combine food with plates to assemble meals. 
Each agent is limited to carrying a single object at a time and cannot directly hand items to other agents: items must be placed on shared counters for indirect transfer. 
We adapt the environment introduced in \cite{wu2021too} to a partially observable setting by adding doors that divide the environment into separate rooms. 
Agents can only observe the contents of the room they currently occupy. 
In this environment, feedback messages can either tell an agent to pass an item to the other agent, or to pick up a different but equivalent item -- so the other agent can take the current one.

\subsection{Models}

\paragraph{\model{}}
For our experiments with mDKG, given that $o^i_t = s_t$, we use a single belief particle ($N=1$). 
As a divergence measure, we compute the probability that the two agents are already acting optimally according to a centralised planner $\pi^*$: $\mathrm{div} = P(\tau \in \pi^* \mid f)$ \cite{ullman2009help}. 
As thresholds, we set $\phi=0$ and $\epsilon = 0.1$.
For Overcooked, to account for partial observability, we use $N=5$ belief particles per agent, and for $\mathrm{div}$ we computed the expected Jensen-Shannon divergence between the action distribution with or without $f$. 
As thresholds, we set $\phi=2$ and $\epsilon = 0.3$.
These values were determined based on a small search on a held-out set (details in the Supplementary).

\begin{figure}[t]
    \centering
    \includegraphics[width=\linewidth]{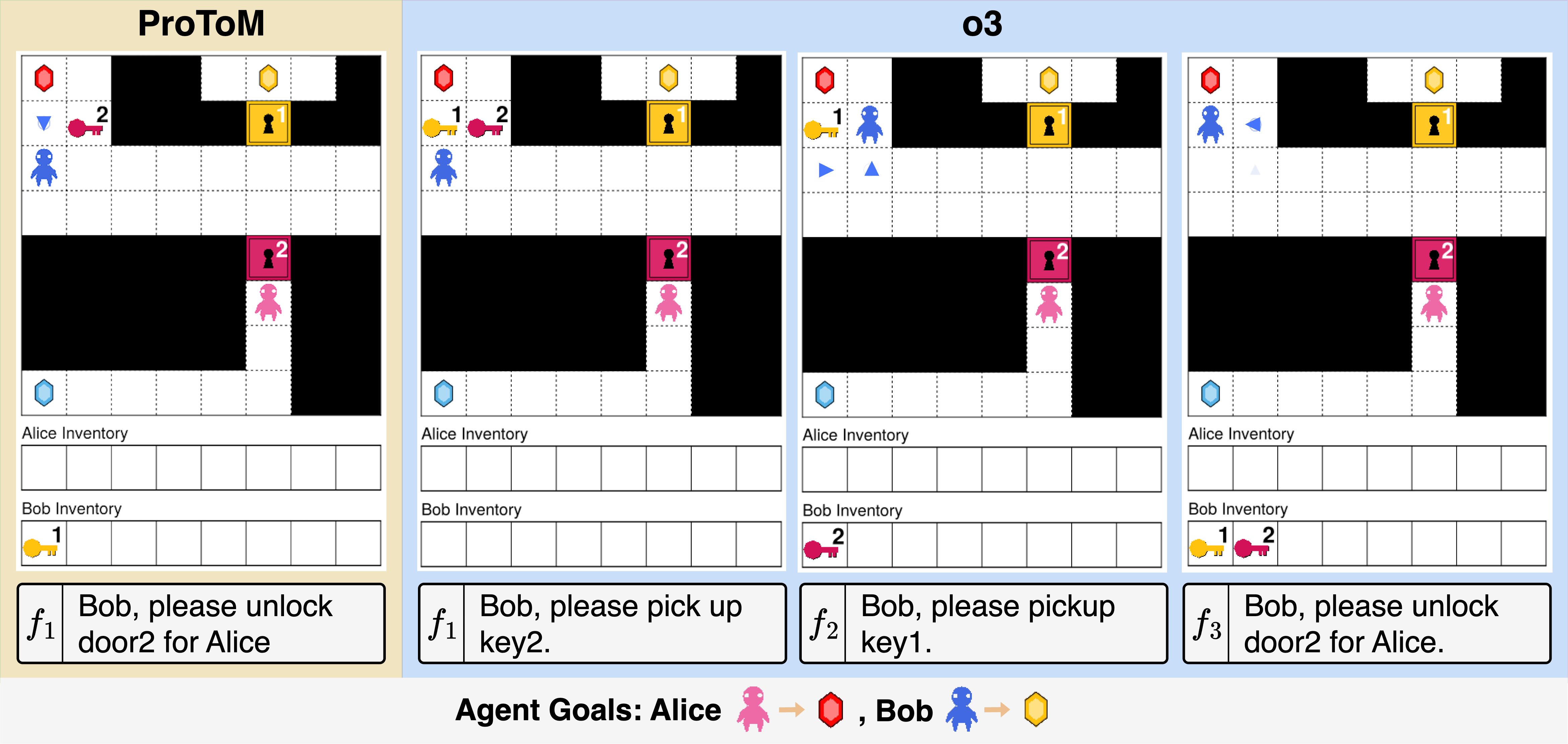}
    \caption{Example comparing \model{} and o3 on mDKG. 
    Both facilitators successfully guide Bob to help Alice by ultimately instructing him to unlock door2. However, while \model{} conveys this key feedback directly, o3 offers overly detailed feedback, resulting in the same outcome but with increased communication overhead.
    }
    \label{fig:mdkg_example}
\end{figure}

\begin{figure}[t]
    \centering
    \includegraphics[width=\linewidth]{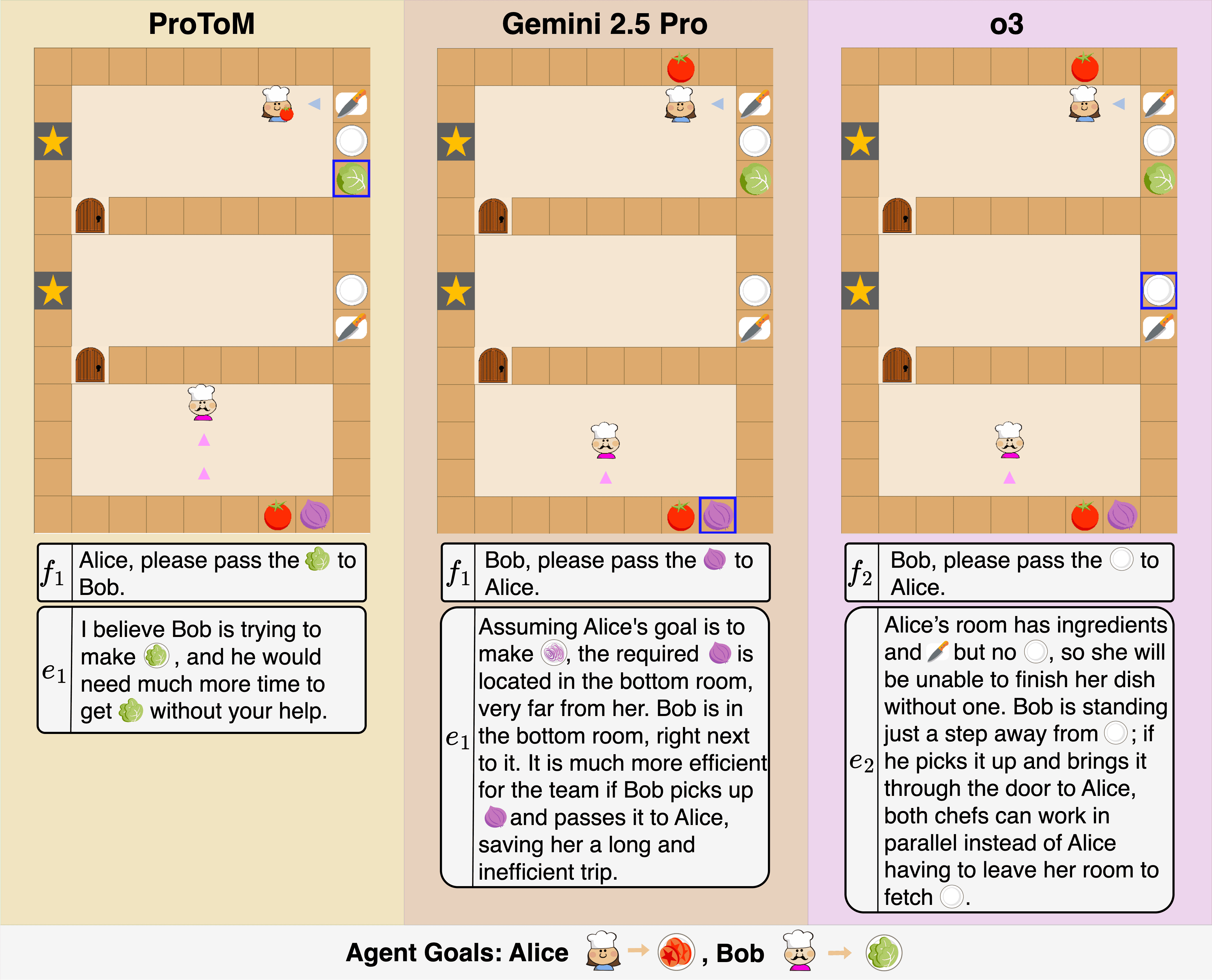}
    \caption{
    Example comparing \model{}, Gemini 2.5 Pro, and o3 on Overcooked. 
    \model{} correctly infers Bob's goal, instructing Alice to pass the ingredient he needs to save him some time. 
    In contrast, o3 and Gemini fail to identify agents' goals, suggesting Bob to pass irrelevant items. 
    In this example, o3 also incorrectly states that Bob is close to a plate, suggesting a poor spatial understanding of the environment. 
    }
    \label{fig:oc_example}
\end{figure}

\begin{figure}[t]
    \centering

    \begin{minipage}{\linewidth}
        \centering
        \subfigure{%
            \includegraphics[width=0.3\linewidth]{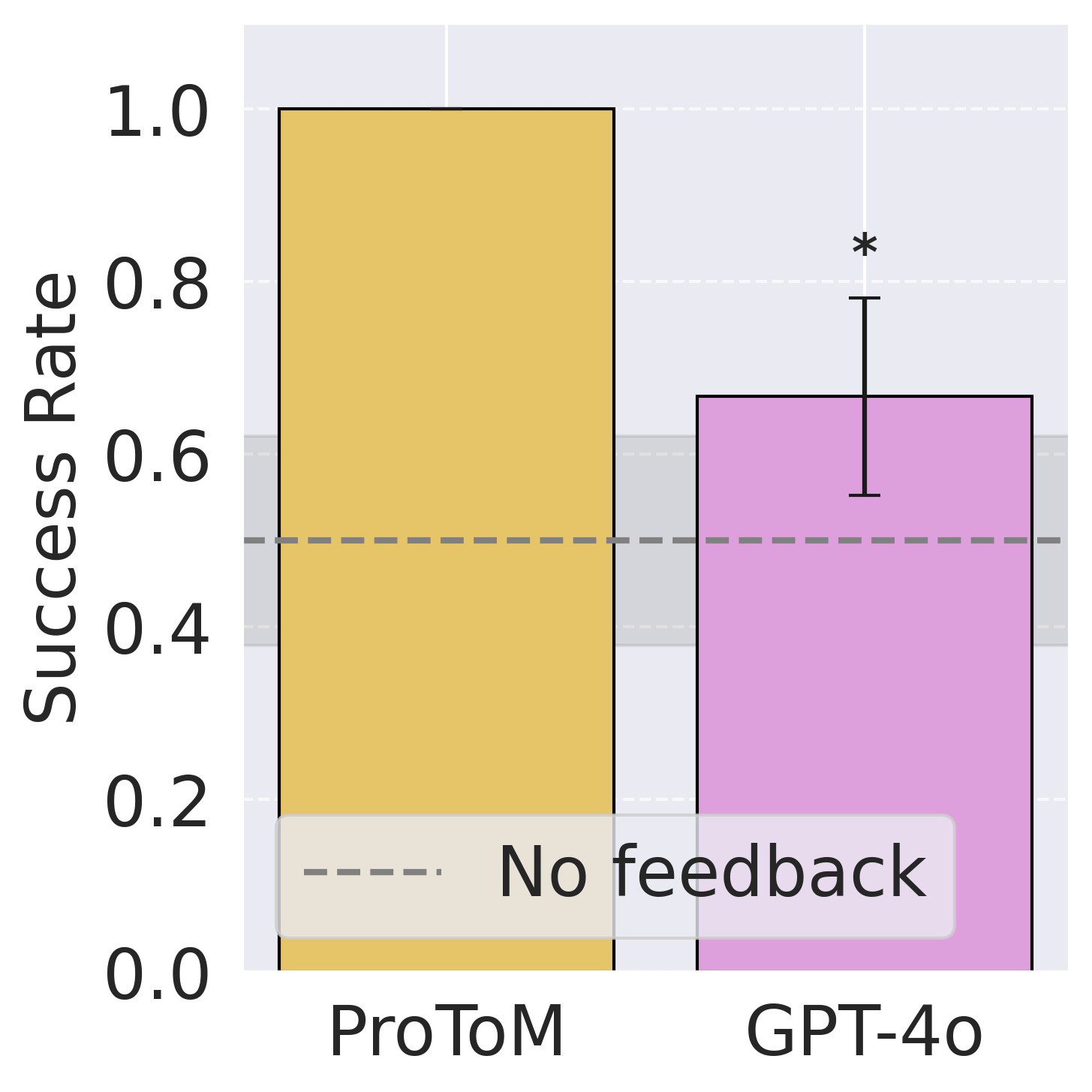}
            \label{fig:human-sr}
        }
        \hfill
        \subfigure{%
            \includegraphics[width=0.3\linewidth]{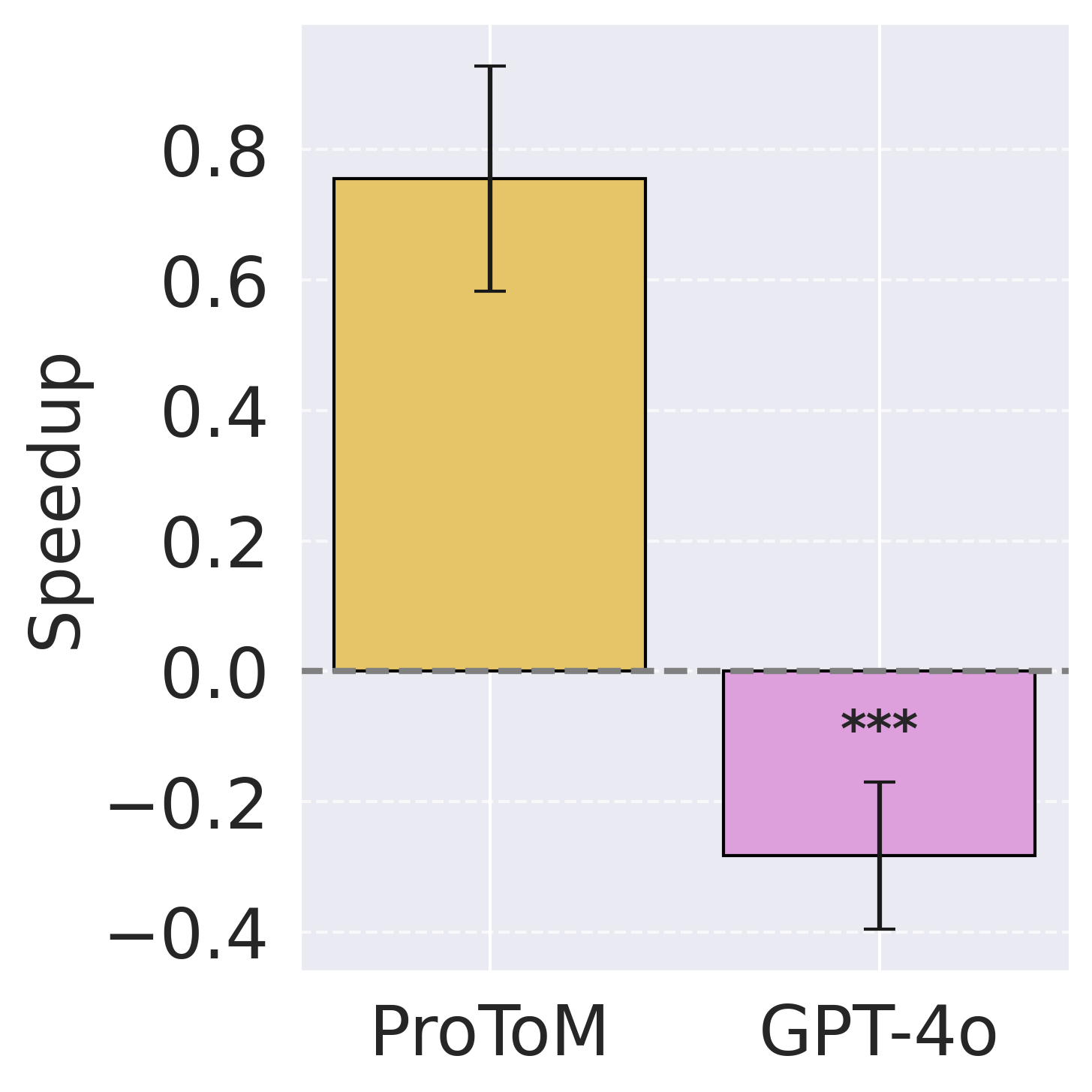}
            \label{fig:human-su}
        }
        \hfill
        \subfigure{%
            \includegraphics[width=0.3\linewidth]{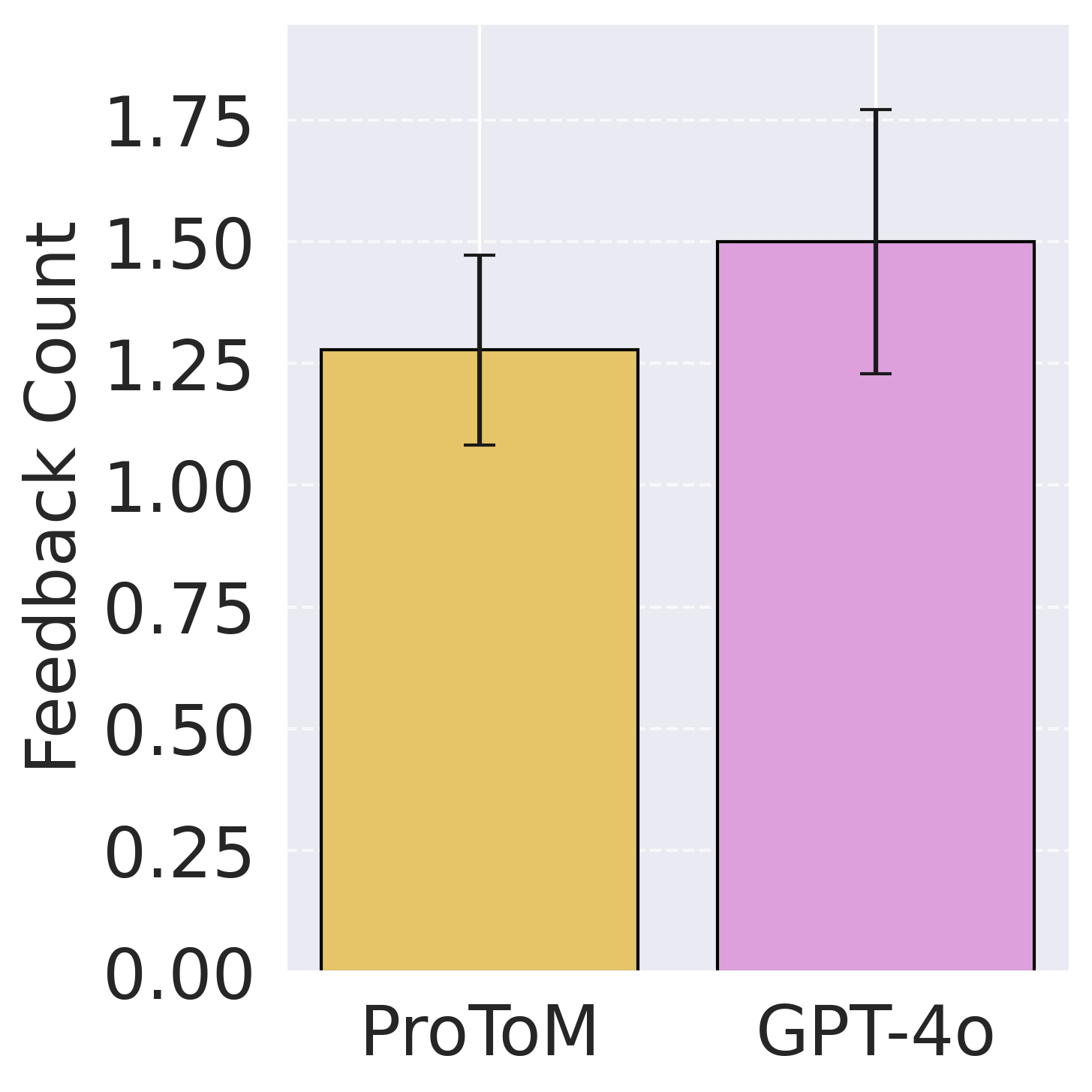}
            \label{fig:human-fc}
        }
    \end{minipage}

    \begin{minipage}{\linewidth}
        \centering
        \includegraphics[width=\linewidth]{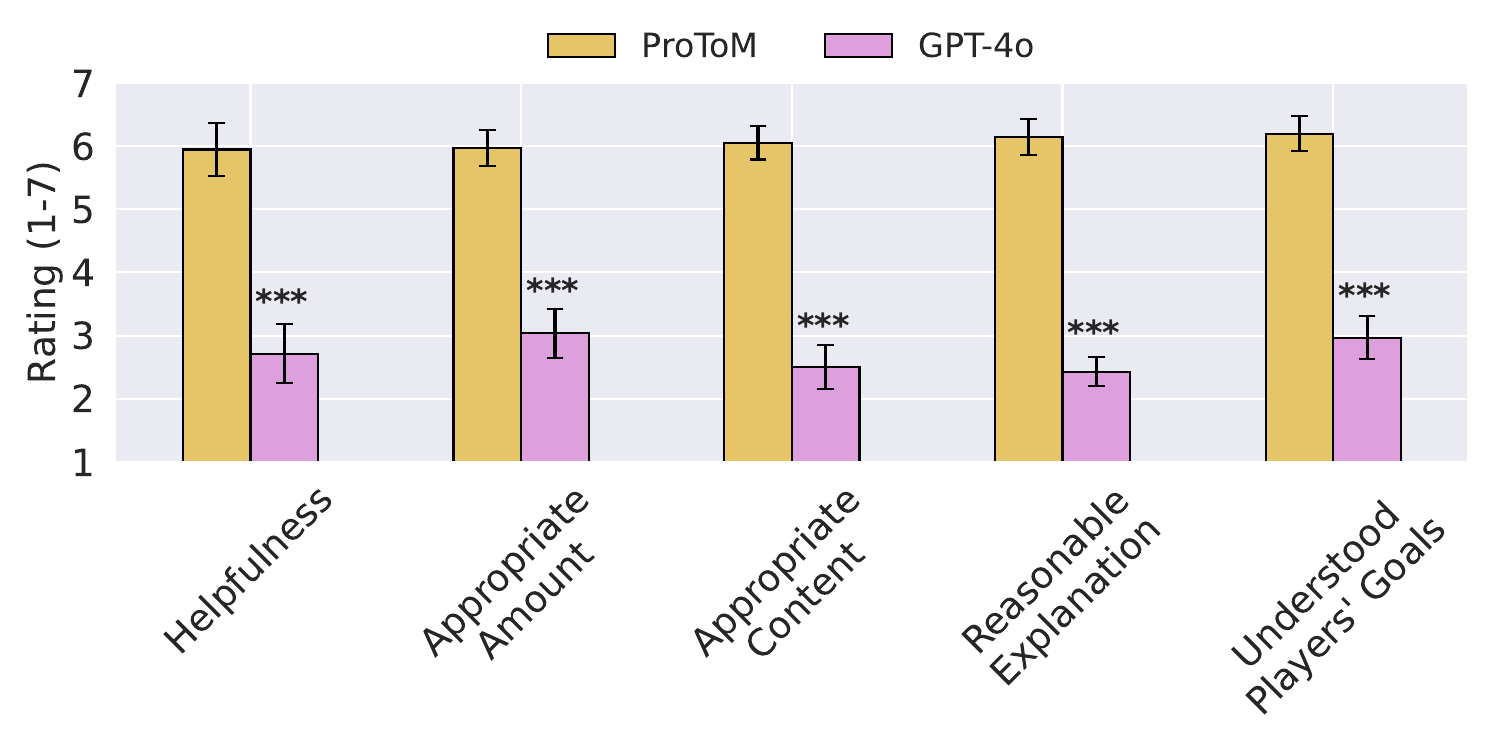}
        \label{fig:human-ratings}
    \end{minipage}

    \caption{In our human study, \model{} significantly outperformed GPT-4o in success rate and task speedup, while requiring similar communication. Participants rated \model{}'s feedback as more helpful, appropriate, better explained and aligned with their goals (*: $p<0.05$, ***: $p<0.001$).}
    \label{fig:human-results}
\end{figure}

\paragraph{Baselines}

We compare \model{} against different baselines:
\textbf{No Facilitator}: Agents receive no feedback.
\textbf{ProToM-Oracle}: Oracle version of \model{} that computes feedback utility (Eq.~\ref{eq:utility}) using agents' ground truth goals.
\textbf{Random Facilitator}: At each timestep, it samples a feedback message uniformly at random from $\mathcal{F}_t$, and decides whether to communicate it with probability $0.5$.
\textbf{Large Vision-Language (VLM) and Reasoning Models (RM)}: Each model is provided with a description of the environment, including object types, agent action and observation space, and task rules. 
In addition, the model is given: an image of the current environment state, a history of recent actions, and the full set of candidate feedback messages $\mathcal{F}_t$. 
The model is then instructed to: (1) infer each agent's goal based on the context, and (2) evaluate whether any message in $\mathcal{F}_t$ would promote prosocial actions that positively impact task efficiency. 
If no feedback is deemed useful, the model is encouraged to respond with \texttt{No Feedback}.
When assisting human participants, we also prompt the model to provide a short explanation of why it choose a specific feedback based on the inferred agent goals.
We evaluate o3, o4-mini \cite{o3-o4mini}, GPT-4o \cite{gpt4o}, Claude 4 Sonnet \cite{claude}, Gemini 2.5 Flash and Pro \cite{comanici2025gemini}, and Qwen 2.5 VL 72B \cite{bai2025qwen2}. 
When possible, we set the model temperature to zero. 
The set of candidate feedback messages $\mathcal{F}_t$ is constructed in the same way for all baselines to ensure a fair and consistent comparison.

\subsection{Metrics}
We use three metrics: success rate, speedup, and number of feedback messages. 
The success rate reflects the proportion of episodes in which both agents successfully complete their tasks.
Speedup measures how much faster agents complete an episode with the help of a facilitator, compared to when they receive no feedback: $\mathrm{Speedup} = L_{\emptyset} / L_f-1$, where $L_f$ and $L_{\emptyset}$ are the average episode length with and without facilitator.
The average number of feedback messages reflects how frequently a facilitator communicates with the agents -- indicating whether it adopts a more parsimonious communication strategy or engages in frequent interaction.

\subsection{Simulation Experiments}
We evaluate \model{} and the baselines with simulated human agents on both mDKG and Overcooked. 
In mDKG, we consider 20 scenarios: six where feedback is not needed, eight where feedback is useful, and six where feedback is necessary.
The two human agents are simulated using the A* planner from \cite{zhi2024pragmatic}. 
In Overcooked, we evaluate 21 scenarios: seven where feedback is not needed, seven where feedback is useful, and seven where feedback is necessary. 
Human agents are simulated using a stochastic heuristic-based planner. 

\paragraph{Results} 
Figure~\ref{fig:sim-results} shows average scores across all episodes. For statistical analysis, we used Fisher's exact test for success rates and the Mann-Whitney U test for other metrics, applying Benjamini-Hochberg correction to control for false positives.

On mDKG, \model{} achieves a perfect success rate ($1.00 \pm 0.00$), and performs on par with \model{}-Oracle, demonstrating effective goal inference.  
Strong reasoning models also achieve high success rates, e.g. $0.95\pm0.05$ both for o3 and Gemini 2.5 Flash. 
However, \model{} achieves a higher average speedup of $0.52 \pm 0.17$ compared to the no-feedback condition, while baselines struggle to improve task efficiency. 
Among the baselines, o3 performs the best with a speedup of $0.16\pm0.14$. 
Notably, \model{} is markedly more selective in its communication, issuing significantly fewer feedback messages than all other models ($0.70\pm0.13$ on average). 
Figure~\ref{fig:mdkg_example} illustrates this qualitatively: while \model{} directly conveys the crucial feedback, o3 offers overly detailed feedback, increasing communication overhead.

Results follow similar trends in Overcooked.
Here, \model{} achieves perfect success rate ($1.00\pm0.00$) and $1.17\pm0.38$ speedup, significantly outperforming all baselines -- with the exception of \model{}-Oracle. 
Despite \model{}'s strong performance, its lower speedup and feedback count compared to \model{}-Oracle -- though not statistically significant -- highlight missed opportunities for useful communication, due to uncertainty about the agent's beliefs and goals in partially observable settings.
Among baselines, Gemini 2.5 Pro achieves the highest success rate ($0.95 \pm 0.05$), followed by o4-mini ($0.90 \pm 0.06$). However both show significantly lower speedups than \model{}:$0.20 \pm 0.20$ and $0.16 \pm 0.20$, respectively.
As in mDKG, \model{} communicates less frequently than all other models ($0.81\pm0.11$ messages on average), with the sole exception of GPT-4o ($1.00\pm0.37$).
We find that VLMs and RMs do not always succeed in inferring agents goals, and therefore often communicate irrelevant feedback. 
Additionally, they sometimes demonstrate poor spatial understanding of the environment.
We show qualitative examples in Figure~\ref{fig:oc_example}.

\subsection{Human Study} 
We conducted a human study in Overcooked with 18 participants grouped into nine pairs, testing three facilitator conditions: \model{}, a VLM, and a control condition with no facilitator. 
Each participant pair played two trials per condition, across six scenarios: three where feedback was useful and three where it was necessary. 
Condition order was randomised to mitigate order effects.
We selected GPT-4o as the LLM facilitator as the best trade-off between performance and response latency.
We found that RMs' longer inference times led to substantial delays at each game timestep ($49$s for o3, on average), which made experiments impractically long (see Supplementary).
Participants were free to follow or ignore the facilitator’s feedback, allowing us to assess both the perceived value and reliability of the assistance.
After each facilitated trial, participants rated the facilitator on four aspects: helpfulness of the feedback, appropriateness of its quantity and content, clarity of explanations, and whether they thought that the facilitator understood their goals. 
Responses were recorded on a 7-point Likert scale (1 = Strongly Disagree, 7 = Strongly Agree).

\paragraph{Results}
Our approach performed best in the human study, as shown in Figure~\ref{fig:human-results}. 
Compared to GPT-4o, \model{} achieved a significantly higher success rate ($1.00 \pm 0.00$ vs $0.66 \pm 0.11$), speedup ($0.75 \pm 0.17$ vs $-0.28 \pm 0.11$), while communicating a similar number of feedback messages ($1.27 \pm 0.19$ vs $1.50 \pm 0.27$).
Notably, GPT-4o's feedback negatively impacted task efficiency, suggesting that it often provided poorly timed or unhelpful feedback.
We also found that human players ignored fewer messages from \model{} than GPT-4o ($5$ vs $10$ in total), again suggesting that \model{}'s feedback was perceived as more relevant, clear, and therefore more trustworthy. 
This is further supported by subjective ratings, shown in Figure~\ref{fig:human-results} (bottom), where participants rated \model{} significantly higher across all criteria.

\subsection{Discussion}
The results across both simulated and human experiments paint a consistent picture: compared to other models, \model{} achieves higher success rates and task speedups, with reduced communication overhead (see Figure~\ref{fig:sim-results} and \ref{fig:human-results}). 
Our human study also showed that \model{}'s feedback was perceived as more helpful, appropriate, and better explained and aligned with users' goals.
By comparison, state‑of‑the‑art VLMs and RMs struggle with communicating feedback that is useful to promote prosocial actions, often causing communication overheads. 
This performance gap in VLMs and RMs is due to two primary limitations of these models. 
First is the lack of strong Theory of Mind abilities: it is well known that existing models struggle to reliably infer mental states \cite{shapira2024clever}.
While modular methods like AutoToM \cite{zhang2025autotom} show promise, their practical application in real-time settings is hindered by the substantial number of model calls required. 
Second, it is well known that current models face challenges in planning \cite{kambhampati2024position}. 
In our setup, these challenges are compounded as planning must incorporate inferred beliefs of multiple agents. 
\section{Related Work}

Theory of Mind, the ability to attribute mental states to oneself and to others \cite{premack1978does}, has been widely studied in collaborative tasks, both using Bayesian \cite{ying2024goma, zhi2024pragmatic}
and neural network approaches \cite{puig2023nopa, ying2024goma, zhang2024proagent, bortoletto24_acl, bortoletto24_ecai, cross2025hypothetical, ruhdorfer2025yokai}. 
Several works focus on enhancing human-AI cooperation, where an AI assistant has to act in the environment to help a single human achieve their goal \cite{ying2024goma, bara2023towards, bortoletto24_acl, puig2021watchandhelp, puig2023nopa, buehler2020theory, yu2024top}.

More closely related to our work, others propose methods to assist human teams, where the AI assistant observes humans acting in the environment and provides feedback to them: 
\citet{seo2023automated} use inverse reinforcement learning to provide task-time interventions that instruct teams on which intent to follow; 
\citet{zhang2024risk} propose a risk-bounded team assistant that, assuming a fixed plan, communicates to prevent failures or resolve deadlocks.
In contrast, our approach supports sequential decision-making, relaxes the assumption that the agents share the same goal, and instead of just avoiding failure focuses on the broader objective of promoting prosocial actions, which can both avoid failure and improve efficiency. 

\section{Conclusion}

In this work, we introduce a novel prosocial facilitator paradigm and \model{}, a method that infers agents’ mental states to select feedback that promotes prosocial behaviour.
\model{} infers agents' goals using Bayesian inverse planning, and selects feedback to communicate by maximising expected utility, conditioned on the inferred goal distribution.
Our evaluations show that while state-of-the-art VLMs and RMs fall short of communicating effective feedback, \model{} provides targeted and helpful feedback, achieving a higher success rate and task speedup -- being consistently preferred by human users.

Our work comes with limitations. 
We have not tested \model{} in real-world settings, which we aim to do in future work. 
Another direction worth exploring further is pragmatic communication, where the facilitator adapts its language based on agents' inferred beliefs and goals. 
Finally, future work could consider more complex settings where the facilitator agents may recursively reason about each other.

\section*{Acknowledgments}
T. Shu acknowledges a grant from Amazon. 

\bibliography{ref}

\clearpage
\appendix
\renewcommand\thefigure{A\arabic{figure}}
\setcounter{figure}{0}
\renewcommand\thetable{A\arabic{table}}
\setcounter{table}{0}
\renewcommand\theequation{A\arabic{equation}}
\setcounter{equation}{0}
\pagenumbering{arabic}
\renewcommand*{\thepage}{A\arabic{page}}
\setcounter{footnote}{0}
\setcounter{page}{1}

\title{Appendix}

\section{Environments}

\subsection{Additional Environment Details}

\paragraph{mDKG} 
\cite{zhi2024pragmatic} is a fully observable, 2D grid-world where the objective is to collect coloured gems, which are often located behind locked doors. 
Unlocking a door requires an agent to possess a key that matches the door's colour. 
Agents can move in the four cardinal directions or remain stationary, while respecting movement constraints imposed by walls and locked doors. 
In addition to navigation, agents can pick up items, hand them over to other agents, or unlock doors when they possess a matching key. 
The possible feedback messages involve either unlocking a door or handing over a key. 
We set the maximum number of timesteps to $T_{max}=80$.

\begin{figure*}[t]
    \centering
    \begin{minipage}{0.25\textwidth}
        \centering
        \subfigure{%
            \includegraphics[width=\linewidth]{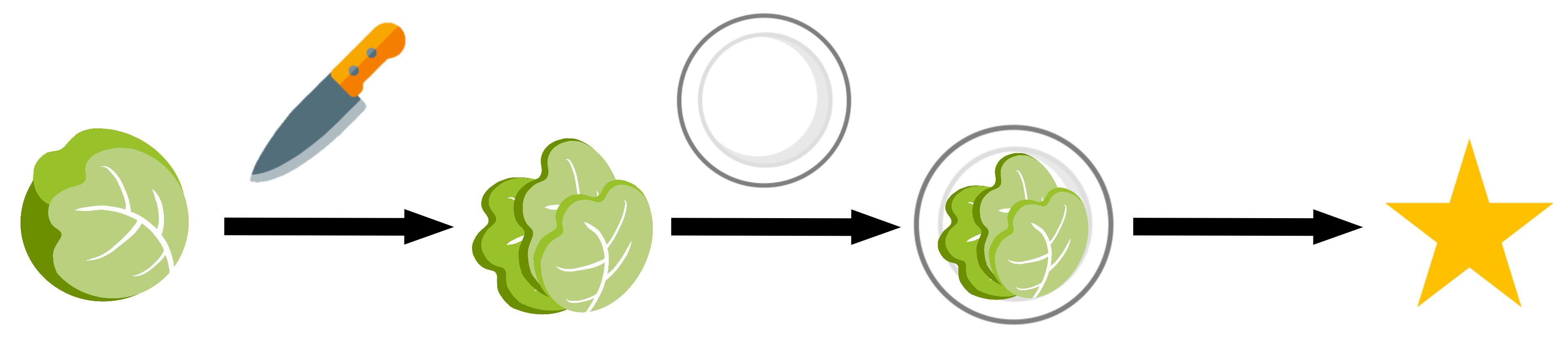}
        }
        \caption{\texttt{SimpleLettuce}}
    \end{minipage}
    \hfill
    \begin{minipage}{0.25\textwidth}
        \centering
        \subfigure{%
            \includegraphics[width=\linewidth]{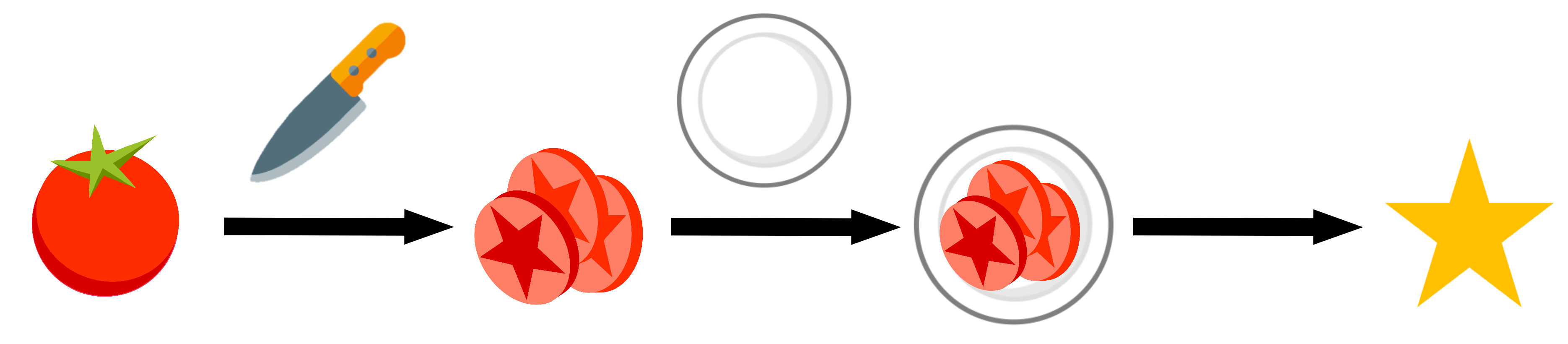}
        }
        \caption{\texttt{SimpleTomato}}
    \end{minipage}
    \hfill
    \begin{minipage}{0.25\textwidth}
        \centering
        \subfigure{%
            \includegraphics[width=\linewidth]{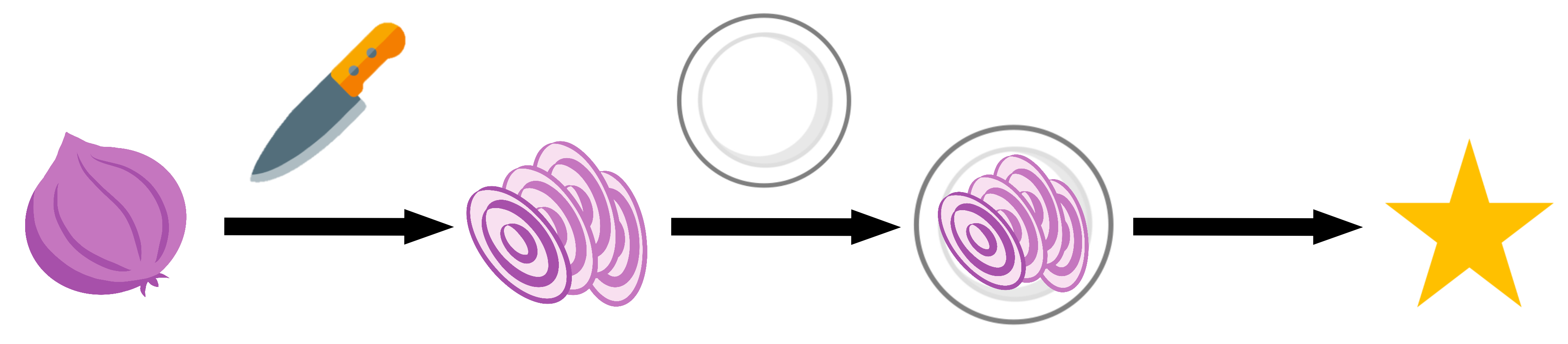}
        }
        \caption{\texttt{SimpleOnion}}
    \end{minipage}
    \caption{
    Possible recipes in Overcooked.
    }
    \label{fig:recipes}
\end{figure*}

\paragraph{Overcooked} 
\cite{carroll2019utility} is a 2D grid-based kitchen in which agents prepare and deliver meals. 
The kitchen consists of counters that can hold either movable items, such as food ingredients and plates, or fixed stations, such as knife stations for chopping and delivery stations for serving completed dishes. 
Agents can move in the four cardinal directions or stay still, pick up and put down items, carry food to knife stations for chopping, and combine food with plates to assemble meals. 
Each agent is limited to carrying a single object at a time and cannot directly hand items to other agents: items must be placed on shared counters for indirect transfer. 
We adapt the environment introduced in \cite{wu2021too} to a partially observable setting by adding doors that divide the environment into separate rooms. 
Agents can only observe the contents of the room they currently occupy. 
The possible recipes to prepare include \texttt{SimpleTomato}, \texttt{SimpleLettuce}, and \texttt{SimpleOnion} -- as shown in Figure~\ref{fig:recipes}. 
Each recipe follows the same three-step process:
\begin{enumerate}
    \item Pick up the main fresh ingredient (e.g., \texttt{FreshTomato}).
    \item Chop it at the chopping station, producing the chopped version (e.g., \texttt{ChoppedTomato}).
    \item Combine the chopped ingredient with a plate (e.g., \texttt{ChoppedTomato-Plate}) and deliver it at the delivery station.
\end{enumerate}
We set the maximum number of timesteps to $T_{max}=100$.

\subsection{Feedback Types}
We define a discrete set of feedback types that capture context-sensitive opportunities to promote prosocial behaviour among agents. These types are instantiated differently in each domain based on their different environment dynamics and available actions.

In mDKG: 
\begin{itemize}
    \item \texttt{Unlock(agent\textsubscript{i}, door\textsubscript{k}, agent\textsubscript{j})}: Suggests agent$_i$ to unlock door$_k$ for agent$_j$. If agent$_i$ doesn't already hold a key to unlock door$_k$, the feedback assumes that agent$_i$ will first pick it up. 
    \item \texttt{Handover(agent\textsubscript{i}, key\textsubscript{k}, agent\textsubscript{j})} Suggests agent$_i$ to hand key$_k$ over to agent$_j$. If agent$_i$ doesn't already key$_k$, the feedback assumes that agent$_i$ will first pick it up. 
\end{itemize}

In Overcooked:
\begin{itemize}
    \item \texttt{Pass(agent\textsubscript{i}, item\textsubscript{k}, agent\textsubscript{j})}: Suggests agent$_i$ to pass a specific item$_k$ at a specified location to another agent across a counter (in Overcooked, agents cannot directly handover items to other agents). Since agents can only hold one item at the time, if agent$_i$ is currently holding an item $\ne$ item$_k$, they are first instructed to place it on the nearest available counter. 
    Then, they pick up the target item and pass it to a shared counter that is not on the layout border and is accessible to both agents.
    \item \texttt{Pickup(agent\textsubscript{i}, item\textsubscript{k})}: Suggests agent$_i$ to pick up a specific item at a specified location. This feedback is primarily used to resolve conflicts where both agents are attempting to pick up the same item, whereas one of them may have access to an alternative. 
\end{itemize}

\begin{figure}[t]
    \centering
    \includegraphics[width=\linewidth]{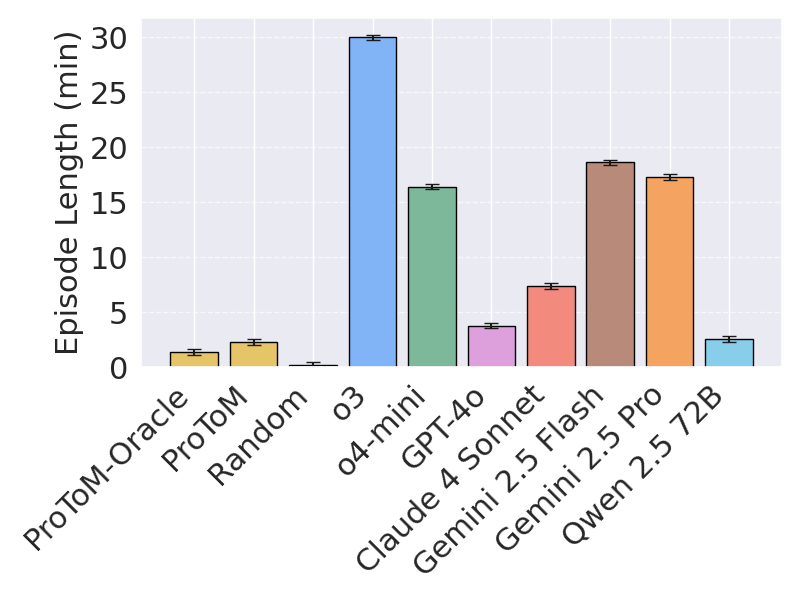}
    \caption{Average episode length in Overcooked, in minutes.}
    \label{fig:runtime}
\end{figure}

\section{Models and Implementation}

\subsection{\model{} Parameters}
For our experiments with mDKG, given that agents have full observability ($o^i_t = s_t$), we use a single belief particle ($N=1$). 
For Overcooked, to account for partial observability, we use $N=5$ belief particles per agent. 
While setting $N=1$ for mDKG is an obvious choice, the number of particles for Overcooked is determined based on the resulting speed of the model inference. 
Given that we use \model{} to assist human agents in in real-time, we opted for $N=5$. 
As thresholds, we set $(\phi=0, \epsilon = 0.1)$ on mDKG, and $(\phi=2, \epsilon = 0.3)$ on Overcooked. 
The values for the thresholds $\phi$ and $\epsilon$ were determined based on a small search on a held-out set, with values from $0$ to $1.0$ increasing by $0.1$ at each search.
For all our experiments, we set the random seed to $42$.

\subsection{\model{} Explanation Templates}
As discussed in the main paper, \model{} generates natural-language explanations for each selected feedback by instantiating predefined templates based on the inferred goal of the other agent and the current environment state. These templates are designed to provide concise and context-aware justifications for the selected feedback.

For the \texttt{Pass(agent\textsubscript{i}, item\textsubscript{k}, agent\textsubscript{j})} feedback, the explanation template is structured as: Template: \texttt{I believe [agent\textsubscript{j}] is trying to prepare [recipe], and...} 
\begin{itemize}
\item If \texttt{agent}\textsubscript{j} can access the item but it would take time to get it: \texttt{... [he/she] would need much more time to get the [item\textsubscript{k}] without your help.}
\item If \texttt{agent}\textsubscript{j} cannot access the item: \texttt{... without your help, [agent\textsubscript{j}\_article] wouldn't be able to get the [item\textsubscript{k}].}
\end{itemize}

For the \texttt{Pickup(agent\textsubscript{i}, item\textsubscript{k})} feedback, the explanation template is:
\texttt{I believe [agent\textsubscript{j}] is trying to prepare [recipe], and the other [item\textsubscript{k}] is easier to get for [him/her].} 

\subsection{LLM Details}
We use the following model versions:
\begin{itemize}
    \item \texttt{o3-2025-04-16} (OpenAI API)
    \item \texttt{o4-mini-2025-04-16} (OpenAI API)
    \item \texttt{gpt-4o-2024-08-06} (OpenAI API)
    \item \texttt{claude-sonnet-4-20250514} (Anthropic API)
    \item \texttt{gemini-2.5-flash} (Google API)
    \item \texttt{gemini-2.5-pro} (Google API)
    \item \texttt{qwen2.5-vl-72b-instruct} (OpenRouter API)
\end{itemize}

\begin{figure}[t]
    \centering
    \includegraphics[width=\linewidth]{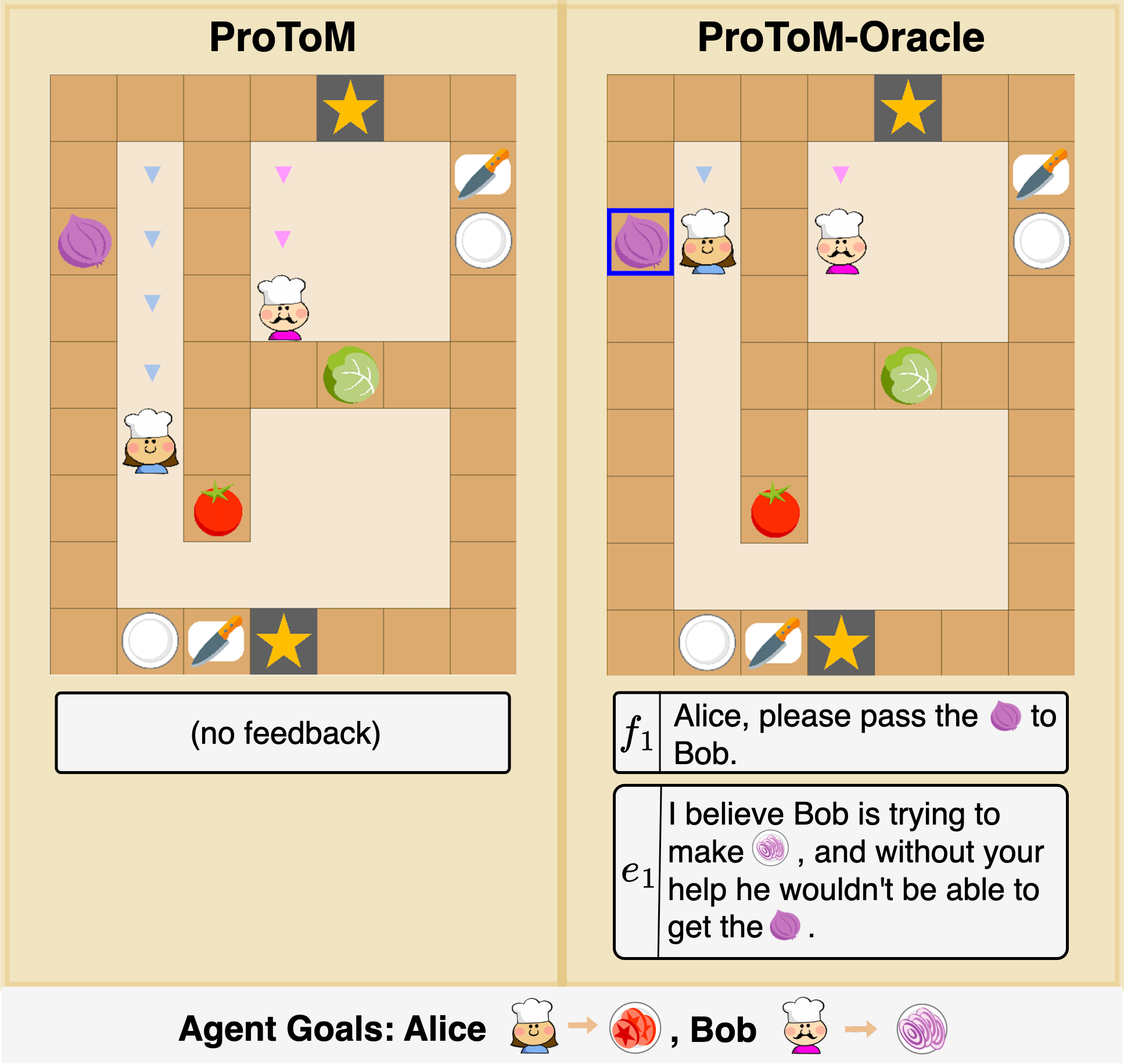}
    \caption{Example in which \model{} fails to communicate feedback, and \model{}-Oracle succeeds.}
    \label{fig:example14}
\end{figure}

\subsection{LLM Prompts}
\begin{figure*}
    \centering
    \includegraphics[width=\linewidth]{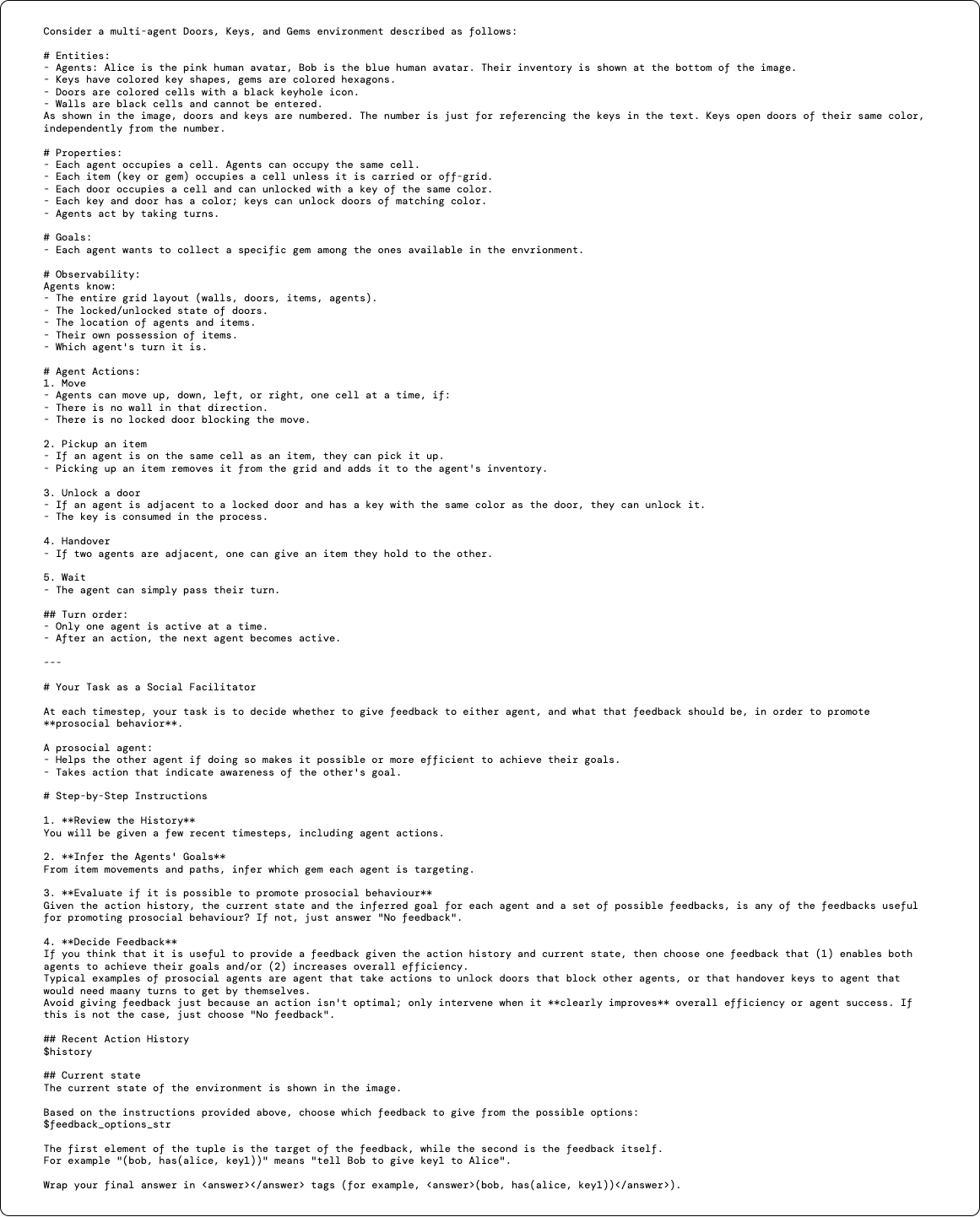}
    \caption{Prompt used for mDKG.}
    \label{fig:mdkg_prompt}
\end{figure*}

\begin{figure*}
    \centering
    \includegraphics[width=\linewidth]{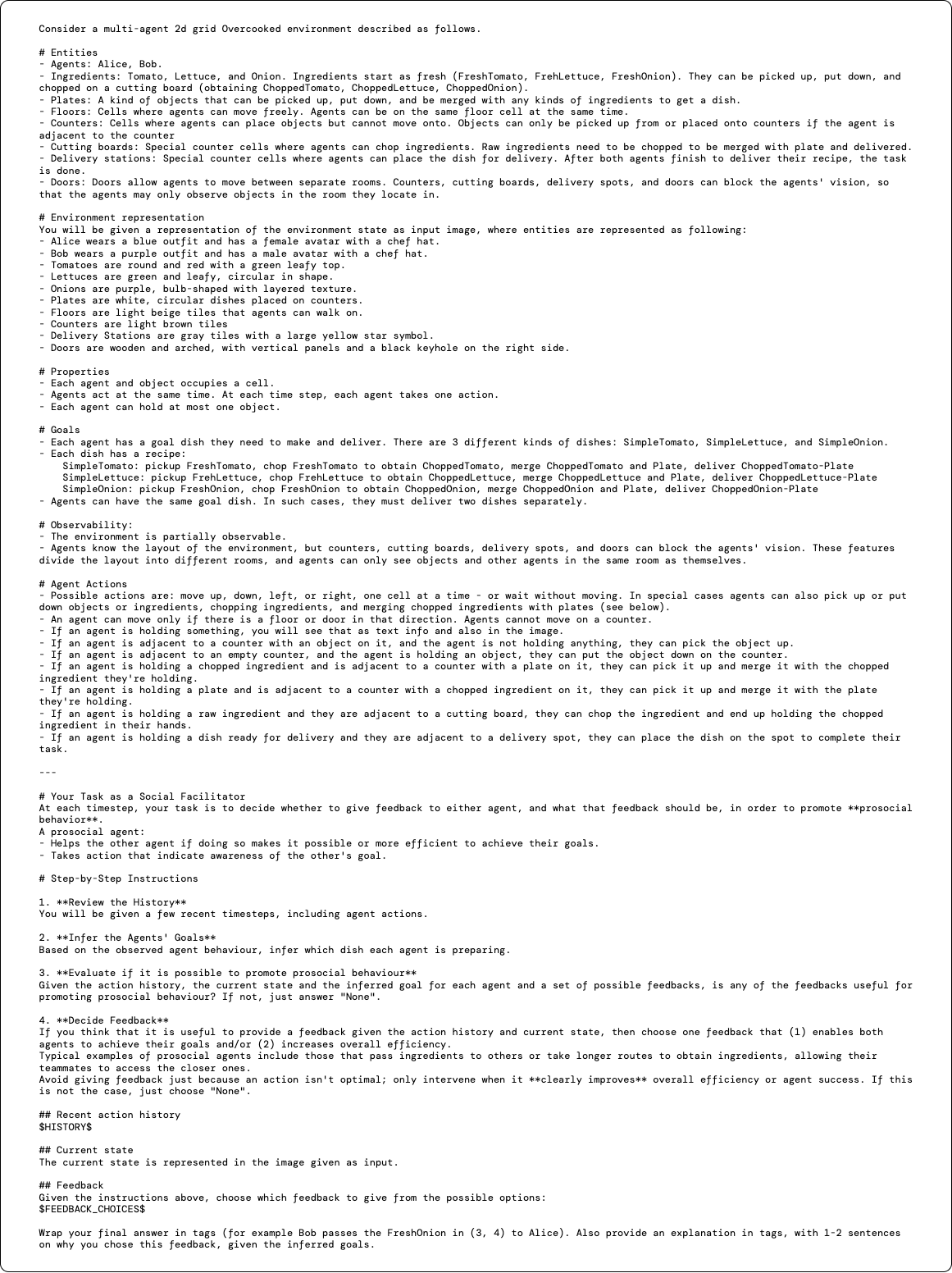}
    \caption{Prompt used for Overcooked.}
    \label{fig:oc_prompt}
\end{figure*}

We show the prompts used for evaluating large VLMs and RMs on mDKG and Overcooked in Figure~\ref{fig:mdkg_prompt} and Figure~\ref{fig:oc_prompt}, respectively.
Each model is provided with a description of the environment, including object types, agent action and observation space, and task rules. 
In addition, the model is given: an image of the current environment state, a history of recent actions, and the full set of candidate feedback messages $\mathcal{F}_t$. 
The model is then instructed to: (1) infer each agent's goal based on the context, and (2) evaluate whether any message in $\mathcal{F}_t$ would promote prosocial actions that positively impact task efficiency. 
If no feedback is deemed useful, the model is encouraged to respond with \texttt{No Feedback}.
When assisting human participants, we also prompt the model to provide a short explanation of why it choose a specific feedback based on the inferred agent goals.
We evaluate o3, o4-mini \cite{o3-o4mini}, GPT-4o \cite{gpt4o}, Claude 4 Sonnet \cite{claude}, Gemini 2.5 Flash, and Gemini 2.5 Pro \cite{comanici2025gemini}. 
When possible, we set the model temperature to zero (RMs often do not allow to control this parameter).

\section{Human Study}

\subsection{Setup and Interface}

\begin{figure*}[t]
    \centering
    \begin{minipage}{0.45\textwidth}
        \centering
        \subfigure{%
            \includegraphics[width=\linewidth]{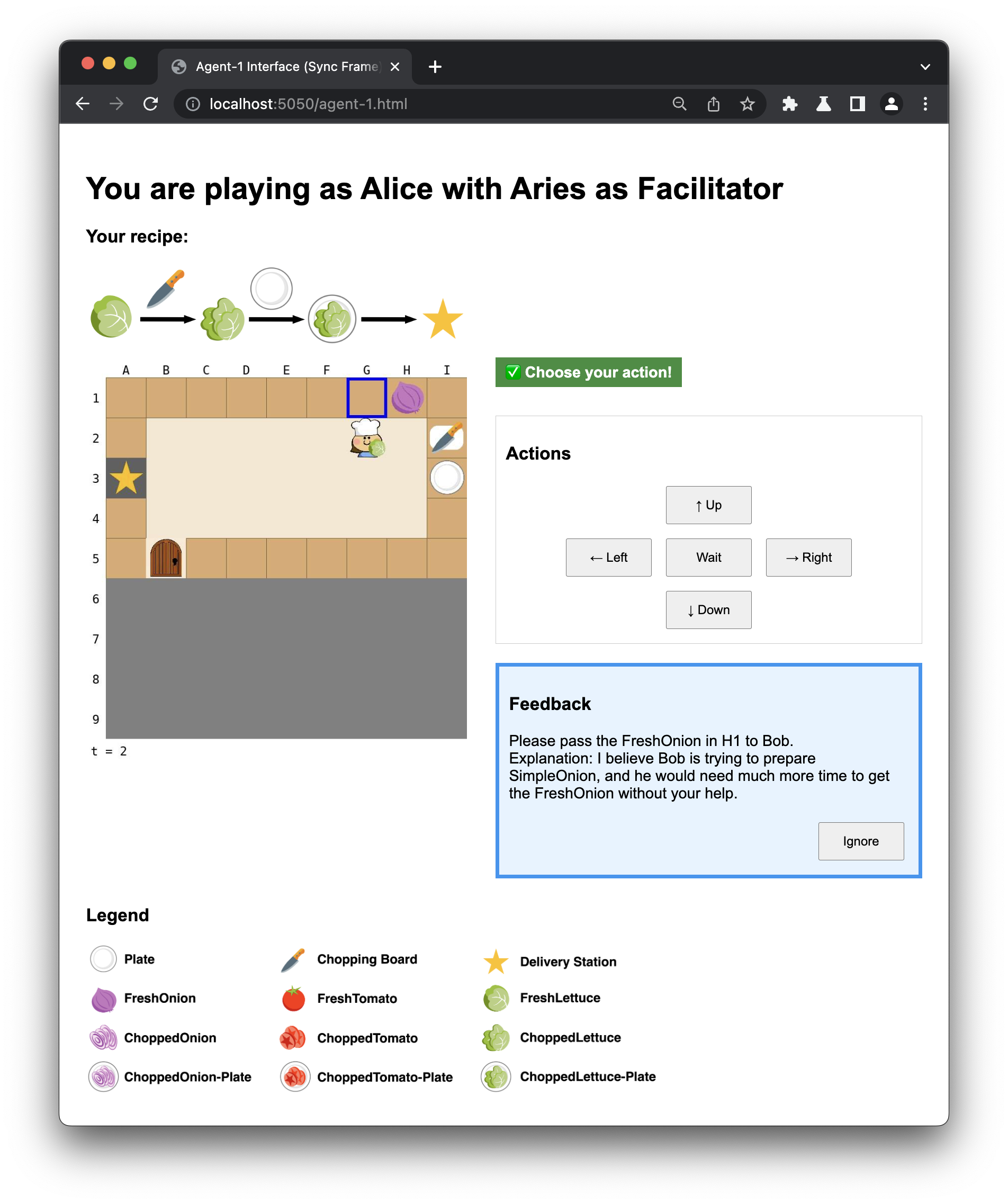}
        }
    \end{minipage}
    \begin{minipage}{0.45\textwidth}
        \centering
        \subfigure{%
            \includegraphics[width=\linewidth]{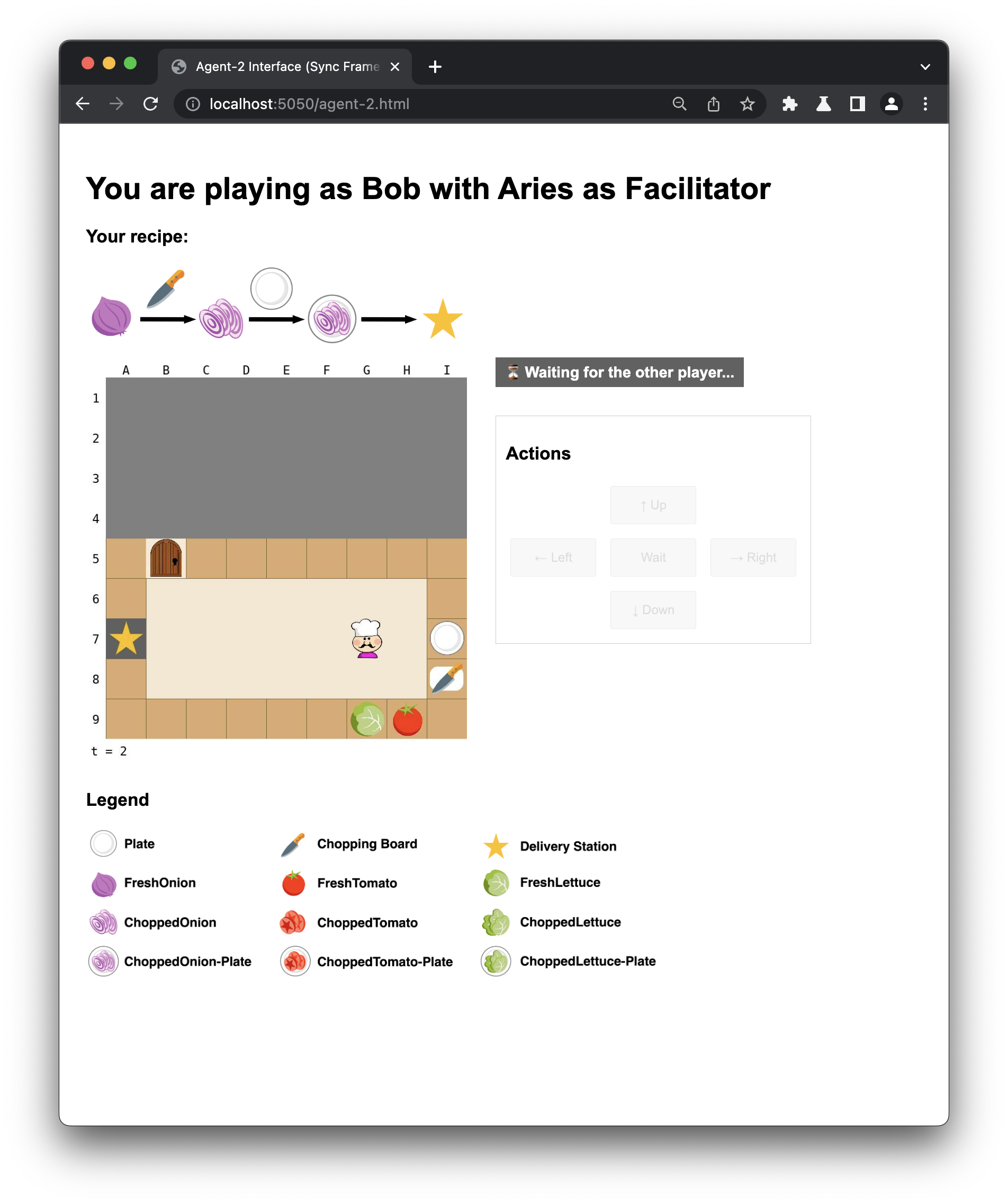}
        }
    \end{minipage}
    \begin{minipage}{0.45\textwidth}
        \centering
        \subfigure{%
            \includegraphics[width=\linewidth]{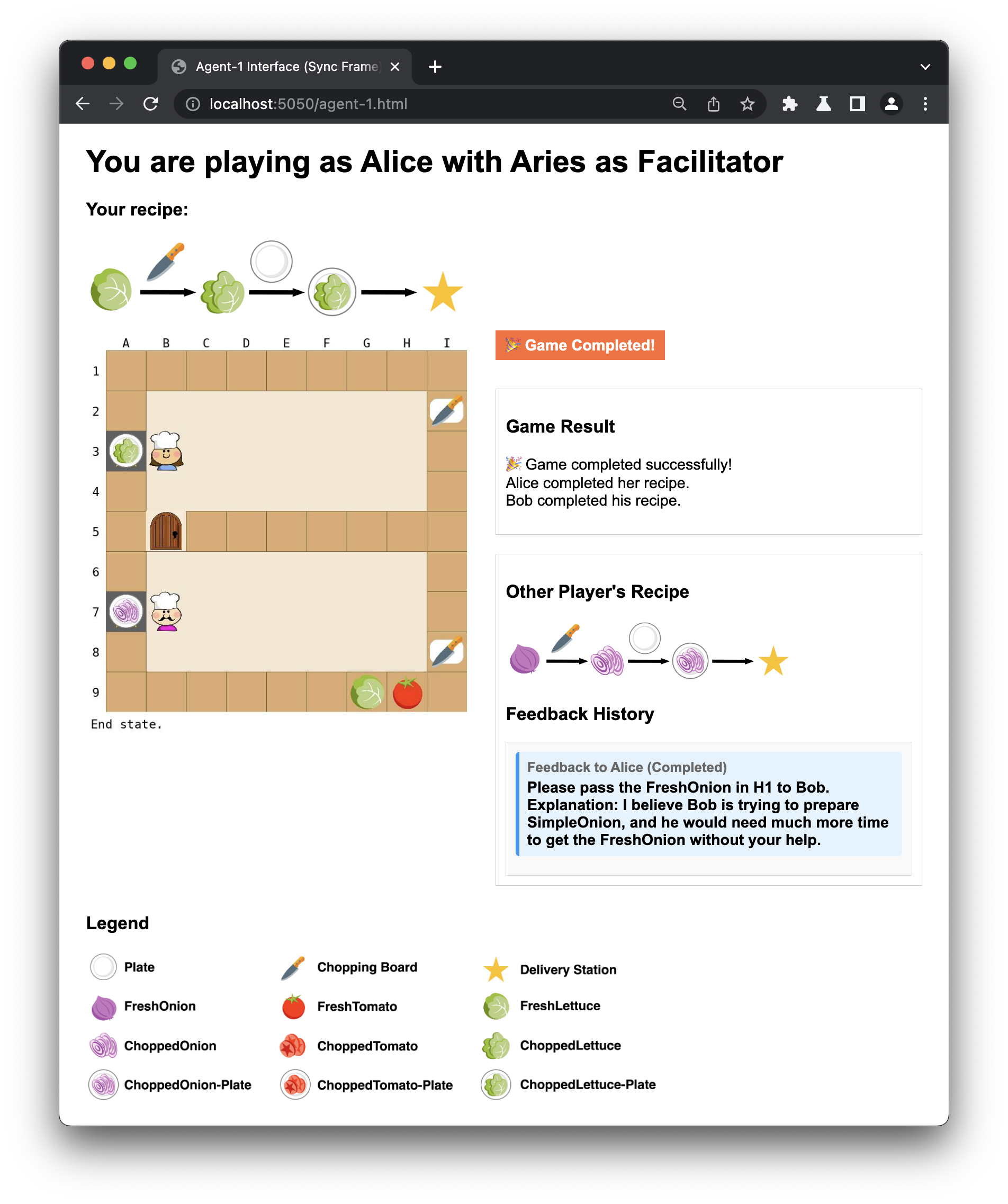}
        }
    \end{minipage}
    \begin{minipage}{0.45\textwidth}
        \centering
        \subfigure{%
            \includegraphics[width=\linewidth]{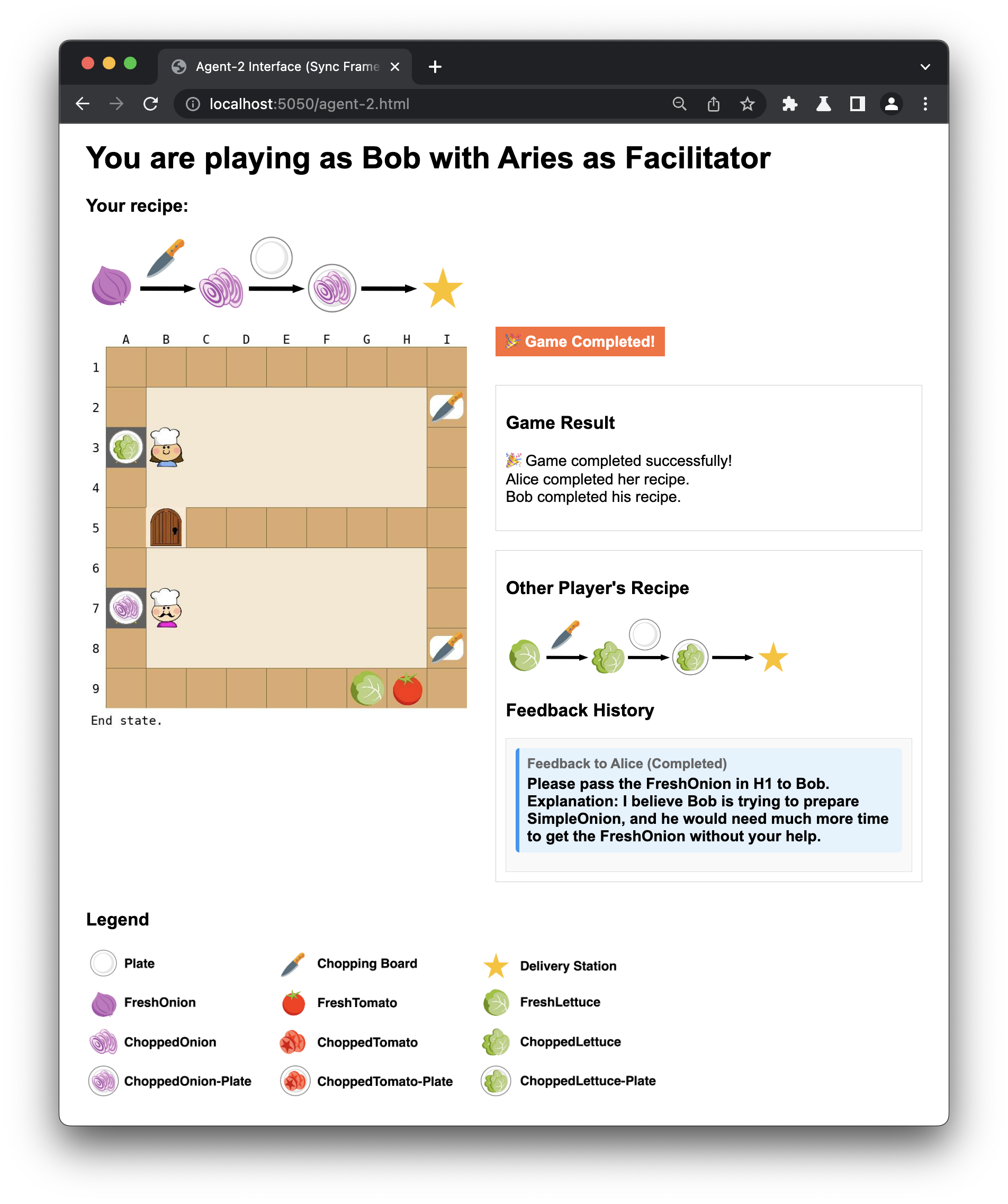}
        }
    \end{minipage}
    \caption{
    Interface for the human study. Each player could see their own goal recipe, their partial observation of the environment, the possible actions to take, and a legend (top). When a player received a feedback message, it would appear as a pop-up blue box, as shown on the left. Players could decide to ignore feedback messages if they considered it useless or unnecessary.
    At the end of each episode, both players were revealed the other player's goal recipe, information about game completion, and the full history of feedback messages given to each player, with their status (``Completed'', ``Ignored'', ``Not Executable'') (bottom). 
    }
    \label{fig:human-study-interface}
\end{figure*}

Figure~\ref{fig:human-study-interface} shows the interface used for the human study. 
Each player could see the codename of the facilitator that was assisting them, their own goal recipe, their partial observation of the environment, the possible actions to take, and a legend. 
When a player received a feedback message, it would appear as a pop-up blue box. 
Players could decide to ignore feedback messages if they considered it useless or unnecessary.
At the end of each episode, both players were revealed the other player's goal recipe, information about episode completion, and the full history of feedback messages given to each player, with their status (``Completed'', ``Ignored'', ``Not Executable''). 
We revealed this information to let both players fill out the questionnaires, even if one of the two players did not receive any feedback message during the episode. 

\subsection{Participant Details and Procedure}
We recruited 18 human participants: 5 female, 13 male, aged between 19 and 31 years old.
The study was approved by the institutional ethics committee.
Some participants were university students who received a compensation, in accordance with university regulations.
The remaining participants voluntarily joined the study, without receiving any form of compensation.
At the beginning of the study, participants were informed about their task, the duration of the experiment, and that their responses would be kept anonymous and used solely for research purposes.
They then went through a guided tutorial that explains the rules of the game.
Each full study -- i.e. two participants playing together the six trials -- lasted around one hour.

\subsection{API Call Times and Justification for Model Choice} 
As we report in the main text, we selected GPT-4o as the LLM facilitator for the human study, given the good trade-off between performance and response latency.
We found that RMs' longer inference times led to substantial delays at each game timestep, which made experiments impractically long. 
We show the average episode length (in minutes) for simulated experiments in Overcooked in Figure~\ref{fig:runtime}. 
As one can see from the figure, models like o3, o4 or Gemini make episode trials drastically longer compared to \model{}. 
Therefore, using one of these models for our human experiments would: 1) be frustrating for the participants; 2) bias or distract them; 3) result in much longer study durations. 
Figure~\ref{fig:runtime} suggests two possible alternatives to GPT-4o: Claude 4 Sonnet and Qwen 2.5 72B. 
However, by looking at Figure~3 (bottom middle) in the main text, we see that both Claude 4 Sonnet and Qwen 2.5 72B engage in lots of feedback communication. 
In preliminary experiments we noticed that such amount of messages would overwhelm the participants, that would tend to just ignore all of them. 
Therefore, we concluded that the best compromise is GPT-4o, which we ended up using in the human study. 

\section{Additional Experimental Results}

\subsection{Quantitative Analysis}
\begin{figure*}[t]
    \centering
    \begin{minipage}{\textwidth}
        \centering
        \subfigure{%
            \includegraphics[width=\linewidth]{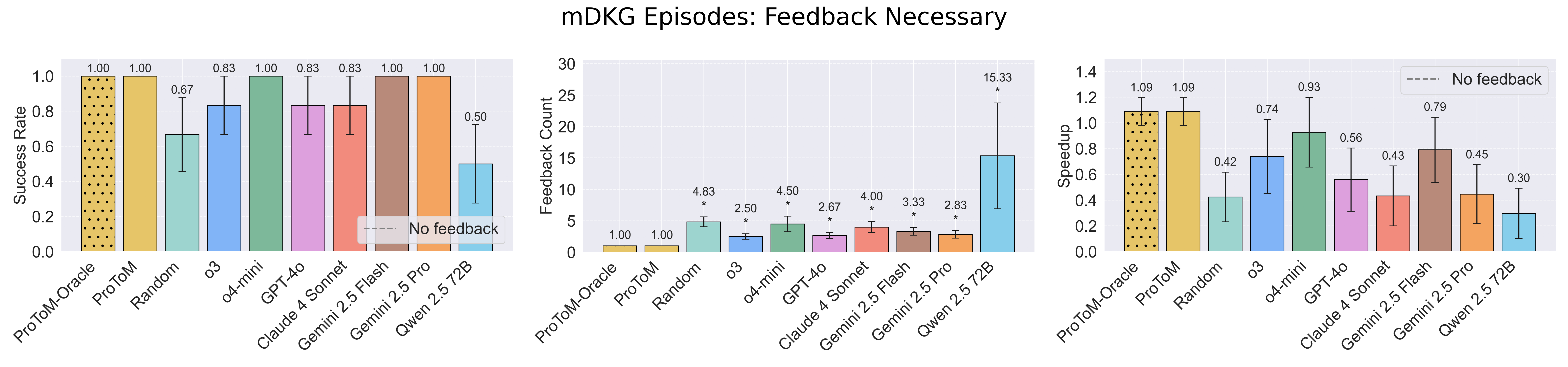}
        }
    \end{minipage}
    \begin{minipage}{\textwidth}
        \centering
        \subfigure{%
            \includegraphics[width=\linewidth]{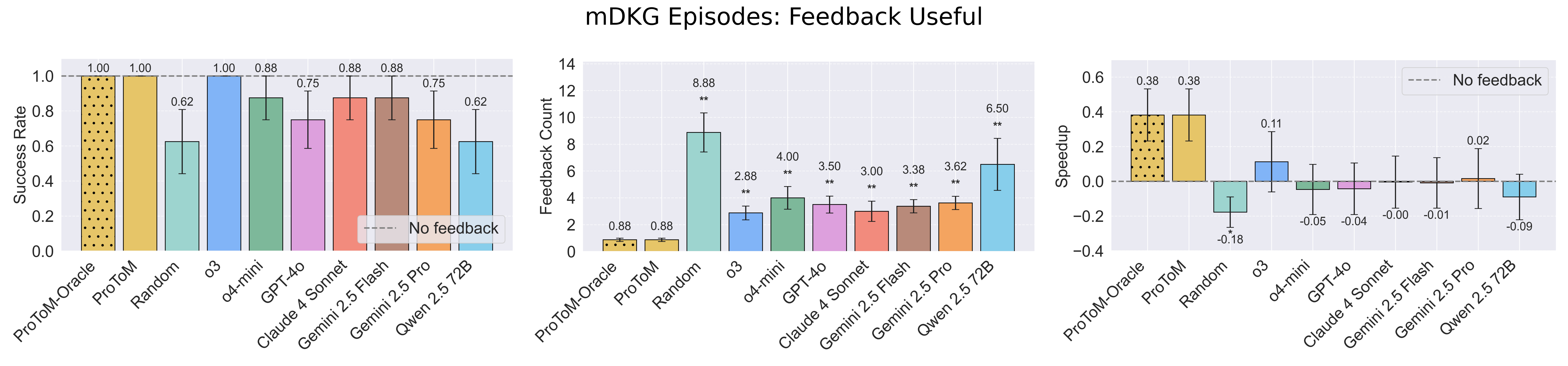}
        }
    \end{minipage}
    \begin{minipage}{\textwidth}
        \centering
        \subfigure{%
            \includegraphics[width=\linewidth]{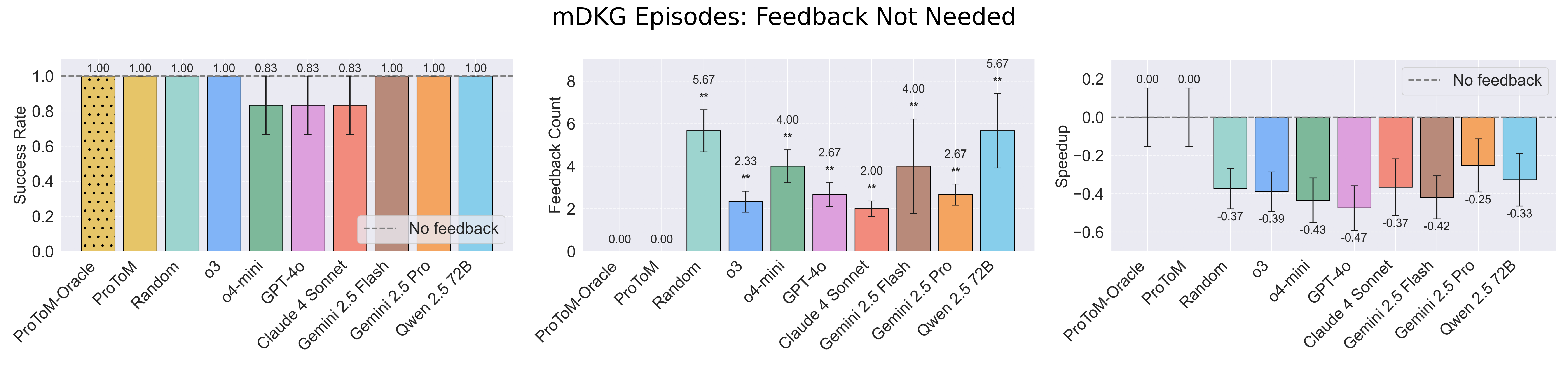}
        }
    \end{minipage}
    \begin{minipage}{\textwidth}
        \centering
        \subfigure{%
            \includegraphics[width=\linewidth]{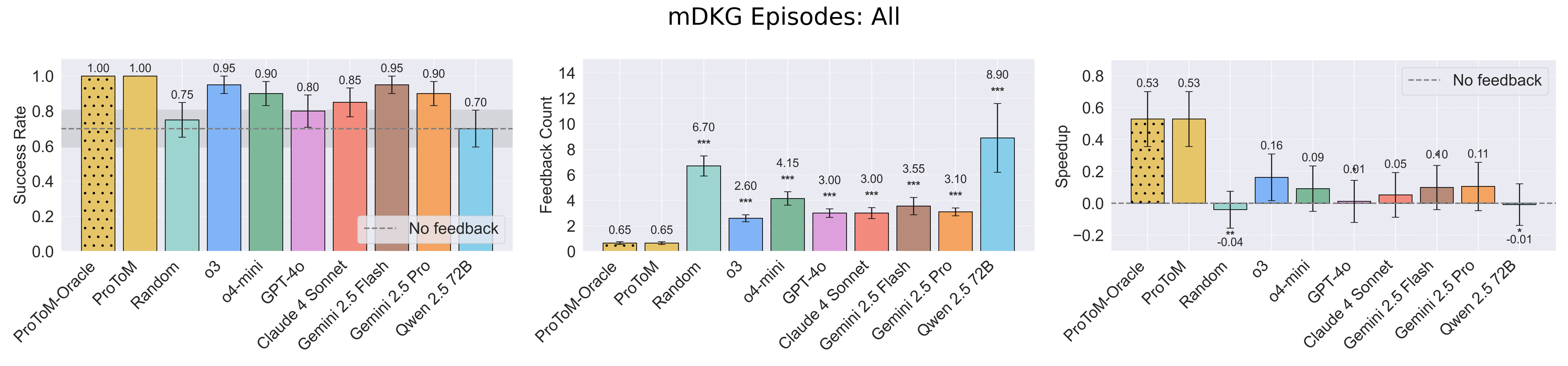}
        }
    \end{minipage}
    \caption{
    for different episode types in the mDKG environment.
    }
    \label{fig:mdkg-types}
\end{figure*}

\begin{figure*}[t]
    \centering
    \begin{minipage}{\textwidth}
        \centering
        \subfigure{%
            \includegraphics[width=\linewidth]{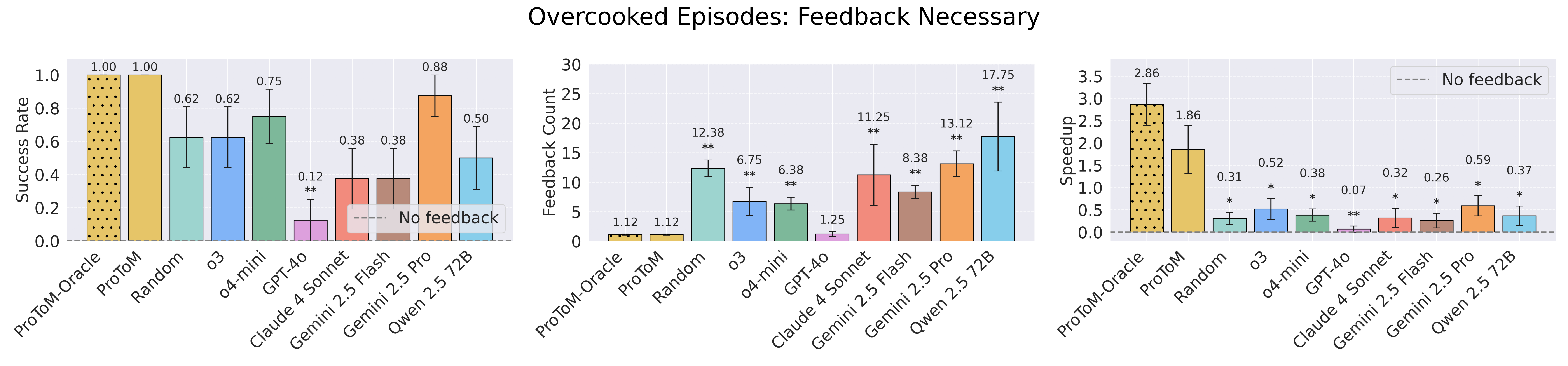}
        }
    \end{minipage}
    \begin{minipage}{\textwidth}
        \centering
        \subfigure{%
            \includegraphics[width=\linewidth]{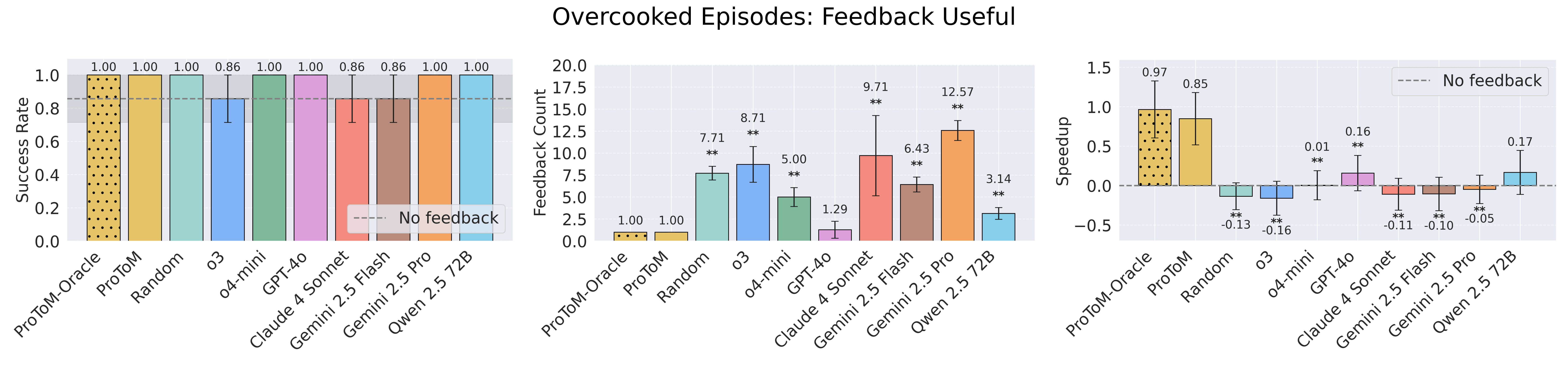}
        }
    \end{minipage}
    \begin{minipage}{\textwidth}
        \centering
        \subfigure{%
            \includegraphics[width=\linewidth]{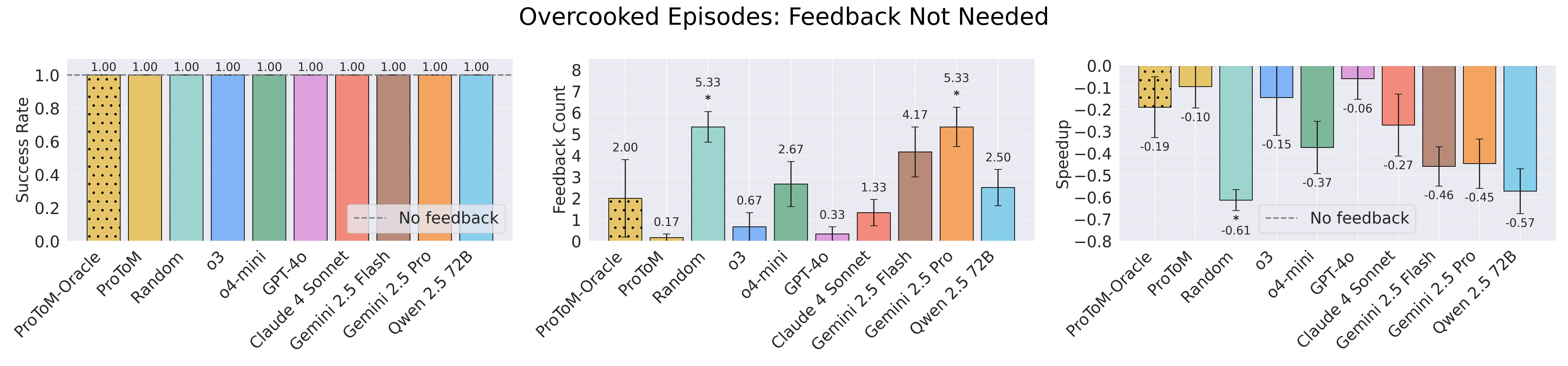}
        }
    \end{minipage}
    \begin{minipage}{\textwidth}
        \centering
        \subfigure{%
            \includegraphics[width=\linewidth]{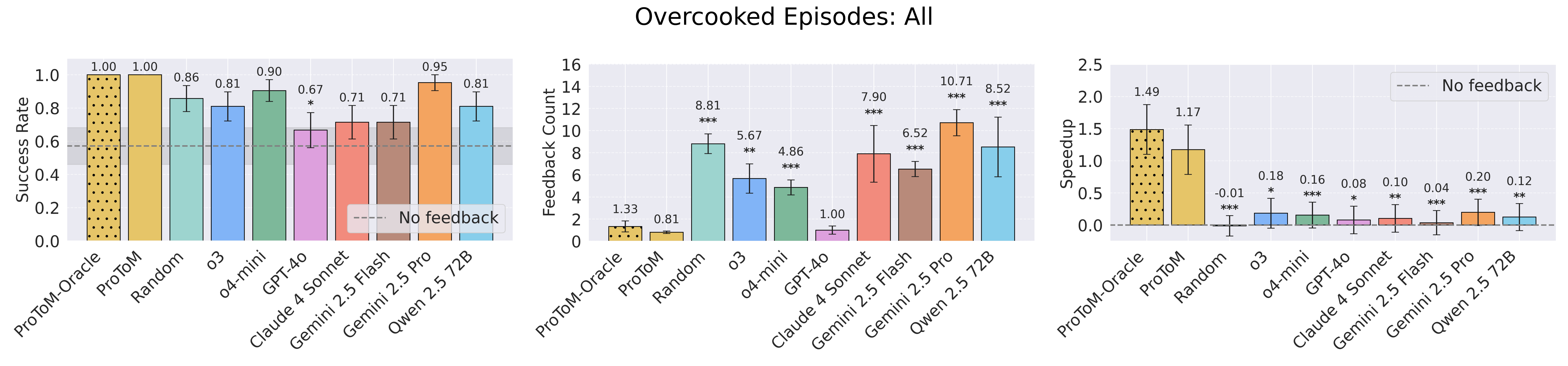}
        }
    \end{minipage}
    \caption{
    Results for different episode types in the Overcooked environment.
    }
    \label{fig:oc-types}
\end{figure*}

For both mDKG and Overcooked, we consider different scenarios where (1) feedback is not needed, as agents can optimally complete their task by themselves; (2) feedback is useful to improve task completion efficiency; (3) feedback is necessary to allow one of the
agents to complete their task. 
While in the main text we provide scores averaged across all scenarios, here we report average scores across each type of scenario in Figure~\ref{fig:mdkg-types} for mDKG, and Figure~\ref{fig:oc-types} for Overcooked.

In mDKG, all as expected: feedback is communicated only in episodes where it is needed or necessary. \model{} is on par with its oracle version (\model{}-Oracle), suggesting strong goal inference. 

In Overcooked, \model{}-Oracle performs better -- although not statistically different -- than \model{} both when feedback is necessary and useful, while communicating the same number of feedback messages (feedback necessary: $1.12 \pm 0.12$; feedback useful: $1.00\pm0.00$). 
This highlights that there are scenarios where \model{} does not communicate, or its feedback is not optimal or is communicated at a non-optimal timestep. 
We show a qualitative example in which \model{} fails in communicating feedback in Figure~\ref{fig:example14}.
On the other hand, when feedback is not needed, \model{}-Oracle communicates more feedback than \model{}, which yields a decrease in task speedup. 
This is due to the fact that, despite knowing the ground truth goals, \model{}-Oracle is still not a perfect model, as there is no ground truth values for the parameters $\phi$ and $\epsilon$. 
For \model{}-Oracle, we used the same values as in \model{}.
We expect that performing a finer grid search on a bigger training set could mitigate this issue. 

\section{Infrastructure and Code}
\subsection{Compute Resources}
We ran our model on a server running Ubuntu 22.04, Intel Xeon Platinum 8260 CPUs for a total of 96 cores. 
Proprietary models are used through API.

\subsection{Code}
Our code is public under the MIT license. 
The code for Overcooked is written in Python 3.10 and adapted from the code released by \cite{wu2021too}. 
The code for mDKG is written in Julia 1.11.5, and based on the Gen.jl \cite{cusumano2019gen} and PDDL.jl \cite{zhi2022pddl} libraries. 

\end{document}